\documentclass{article}

\usepackage{epsfig}
\usepackage{amssymb}
\usepackage{natbib}

\usepackage[utf8]{inputenc} 
\usepackage[T1]{fontenc}    
\usepackage{hyperref}       
\usepackage{url}            
\usepackage{booktabs}       
\usepackage{amsfonts}       
\usepackage{nicefrac}       
\usepackage{microtype}      
\usepackage{xcolor}         

\usepackage{graphicx}
\usepackage{amsmath}
\usepackage{amsthm}

\usepackage{settings_preprint}
\usepackage{authblk}

\author[1]{Lucas De Lara}
\author[1]{Alberto Gonz\'alez-Sanz}
\author[2]{Nicholas Asher}
\author[1]{Laurent Risser}
\author[1]{Jean-Michel Loubes}
\affil[1]{Institut de Mathématiques de Toulouse, Université Paul Sabatier}
\affil[2]{Institut de Recherche en Informatique de Toulouse, Université Paul Sabatier}

\title{Transport-based Counterfactual Models}
\date{}

\begin{document}

\maketitle

\begin{abstract}
  Counterfactual frameworks have grown popular in machine learning for both explaining algorithmic decisions but also defining individual notions of fairness, more intuitive than typical group fairness conditions.
  However, state-of-the-art models to compute counterfactuals are either unrealistic or unfeasible. In particular, while Pearl's causal inference provides appealing rules to calculate counterfactuals, it relies on a model that is unknown and hard to discover in practice. We address the problem of designing realistic and feasible counterfactuals in the absence of a causal model. We define transport-based counterfactual models as collections of joint probability distributions between observable distributions, and show their connection to causal counterfactuals. More specifically, we argue that optimal-transport theory defines relevant transport-based counterfactual models, as they are numerically feasible, statistically-faithful, and can coincide under some assumptions with causal counterfactual models. Finally, these models make counterfactual approaches to fairness feasible, and we illustrate their practicality and efficiency on fair learning. With this paper, we aim at laying out the theoretical foundations for a new, implementable approach to counterfactual thinking.
\end{abstract}

\textbf{Keywords:} Counterfactuals, Optimal transport, Causality,  Fairness, Supervised learning


\section{Introduction} \label{introduction}

A \emph{counterfactual} states how the world should be modified so that a given outcome occurs. For instance, the statement \emph{had you been a woman, you would have gotten half your salary} is a counterfactual relating the \emph{intervention} \say{had you been a woman} to the \emph{outcome} \say{you would have gotten half your salary}. Counterfactuals have been used to define causation \citep{lewis1973} and hence have attracted attention in the fields of explainability and robustness in machine learning, as such statements are tailored to explain black-box decision rules. Applications abound, including algorithmic recourse \citep{joshi2019towards,poyiadzi2020face,karimi2021algorithmic, rasouli2021care,slack2021counterfactual,bajaj2021robust}, defense against adversarial attacks \citep{ribeiro2016should,moosavi2016deepfool} and fairness \citep{kusner2017counterfactual, black2020fliptest, plecko2020fair, asher2021fair}.

State-of-the-art models for computing meaningful counterfactuals have mostly focused on the \emph{nearest counterfactual explanation} principle \citep{wachter2017counterfactual}, according to which one finds minimal translations, minimal changes in the features of an instance that lead to a desired outcome. However, as noted by \cite{black2020fliptest} and \cite{poyiadzi2020face}, this simple distance approach generally fails to describe realistic alternative worlds, as it implicitly assumes the features to be independent. Changing just the gender of a person in such a translation might convert from a typical male into an untypical female, rendering out-of-distribution counterfactuals like the following: {\em if I were a woman I would be 190cm tall and weigh 85 kg}. According to intuition, such counterfactuals are false and rightly so because they are not representative of the underlying statistical distributions. As a practical consequence, such counterfactuals typically hide biases in machine learning decision rules \citep{lipston2018does,besse2021survey}.

The link between counterfactual modality and causality motivated the use of Pearl's causal modeling \citep{pearl2009causality} to address the aforementioned shortcoming \citep{kusner2017counterfactual,joshi2019towards,mahajan2020preserving,karimi2021algorithmic}. Pearl's do-calculus, by enforcing a change in a set of variables while keeping the rest of the causal mechanism untouched, provides a rigorous basis for generating intuitively true counterfactuals. The cost of this approach is fully specifying the causal model, namely specifying not only the Bayesian network (or graph) capturing the causal links between variables, but also the structural equations relating them, and the law of the latent, exogenous variables. The reliance on such a strong prior makes the causal approach appealing in theory, but inadequate for deployment on practical cases.

To sum-up, research has mostly focused on two divergent frameworks to compute counterfactuals: one that proposes an easy-to-implement model that leads, however, to intuitively untrue counterfactuals; another rigorously takes into account the dependencies between variables to produce realistic counterfactuals, but at the cost of feasibility. Our contribution addresses a third way. Extending the work of \cite{black2020fliptest}, who first suggested substituting causality-based counterfactual reasoning with optimal transport, we define \emph{transport-based counterfactual models}. Such models, by characterizing a counterfactual operation as a coupling, a mass transportation plan between two observable distributions, ensures that the generated counterfactuals are in-distribution, hence realistic. In addition, they remedy to the impracticability issues of causal modeling as they can be computed through any mass transportation techniques, for instance optimal transport. The major benefit of this approach is that it renders doable many critical applications of counterfactual frameworks, for example in algorithmic fairness.

\subsection{Outline of contributions}

We make both theoretical and practical contributions in the fields of counterfactual reasoning and fair machine learning. We propose a mass-transportation framework for counterfactual reasoning and point out its similarities to the causal approach. Additionally, we show that counterfactual methods for fairness become feasible in this framework by introducing and implementing transport-based counterfactual fairness criteria. More precisely, our contributions can be outlined as follows.  

\begin{enumerate}
    \item In Section \ref{sec:causal}, we recall the necessary background on Pearl's causal modeling, while we introduce in Section \ref{sec:transport} the basics of mass transportation and optimal transport theory. Both sections serve as the theoretical and notational toolbox that will be used throughout; they are meant to keep the paper self-contained and can be skipped by readers familiar with these subjects.
    \item In Section \ref{sec:models}, we firstly recall how to compute counterfactual quantities using causal modeling. Then, we introduce a general causality-free framework for the computation of counterfactuals through mass-transportation techniques, encompassing the approach of \cite{black2020fliptest}. Essentially, we also propose a unified mass-transportation viewpoint of counterfactuals, be them causal-based or transport-based, through the definition \emph{counterfactual models}, collections of couplings characterizing all possible counterfactual statements for a given feature to alter (for example the gender). We provide concrete examples of models, and discuss the limitations of the different approaches.
    \item In Section \ref{sec:revisit}, we leverage the unified formalism proposed in the previous section to demonstrate connections between causality and optimal transport. More precisely, after studying the implications of two general causal assumptions onto the induced counterfactual models, we demonstrate that optimal transport maps for the quadratic cost generates the same counterfactual instances as some specific causal models, including the common linear additive models. We argue that this makes optimal-transport-based counterfactual models relevant surrogates in the absence of a known causal model.
    \item In Sections \ref{sec:fairness}, \ref{sec:application} and \ref{sec:numerical}, we illustrate the practicality of our approach for fairness in machine learning. We apply the mass-transportation viewpoint of structural counterfactuals by recasting the \emph{counterfactual fairness} criterion \citep{kusner2017counterfactual} into a transport-like one. Then, we propose new causality-free criteria by substituting the causal model by transport-based models in the original criterion. Finally, we address the training of counterfactually fair classifiers, providing statistical guarantees and numerical experiments over various datasets.
\end{enumerate}

To sum-up: Sections~\ref{sec:causal}~and~\ref{sec:transport} provide the prerequisites for the paper; Sections~\ref{sec:models}~and~\ref{sec:revisit} introduce the concept of counterfactual models and the corresponding theory; Sections~\ref{sec:fairness}~to~\ref{sec:numerical} address fairness applications of these models.

\subsection{Related work}

This work follows the paper of \cite{black2020fliptest}, which focus on building sound counterfactual quantities through optimal transport, deviating from both causal-based techniques and the nearest-counterfactual-instance principle. Our contributions in Sections~\ref{sec:models}~and~\ref{sec:revisit} can be seen as the theoretical foundations of their approach, by shedding light on the link between measure-preserving counterfactuals and structural counterfactuals. Also, we note that the way we introduce the causal account for counterfactual reasoning in Section~\ref{sec:models} concurs with \citep{plecko2020fair} and \citep{bongers2021foundations}. More precisely, we underline that the objects of interest are the joint probability distributions, or couplings, generated by manipulations of the causal model. Additionally, we propose in Section~\ref{sec:fairness} a direct extension of the counterfactual fairness frameworks introduced in \citep{kusner2017counterfactual} and \citep{russell2017when} to transport-based counterfactual models, leading to a new method for supervised fair learning. This relates our work to the rich literature on fair learning through optimal transport \citep{gordaliza2019obtaining, chiappa2020general, gouic2020projection, evgenii2020fair, risser2022tackling}. Finally, we note that the main result of Section~\ref{sec:revisit}, stating that optimal transport maps recover causal effects under specific assumptions, shares similarities with the main theorem of \citep{torous2021optimal}. In contrast to our work, their assumptions are motivated by the study of heterogeneous treatment effects, which concerns counterfactual inference in the Neyman-Rubin causal framework \citep{rubin1974estimating, imbens2015causal}.

\section{Causal modeling}\label{sec:causal}

Pearl's causal modeling addresses the fundamental problem of analyzing causal relations between variables, beyond mere correlations \citep{pearl2009causality}. It can be regarded as a mathematical formalism meant to describe associations that standard probability calculus cannot \citep{pearl2010mathematics}. This section recalls the basic theory on this modeling, borrowing the rigorous mathematical framework recently proposed by \cite{bongers2021foundations}. It is meant to keep the paper self-contained and can be skipped by a reader familiar with causality.

Let us fix some notations before proceeding. Throughout the paper, we consider a probability space $(\Omega, \mathcal{A}, \P)$. We denote respectively by $\mathcal{L}(W)$ and $\E[W]$ the law and expectation under $\P$ of any random variable $W$ on $\Omega$ taking values in a measurable space $(\R^p,\mathcal{B})$ where $p \geq 1$ and $\mathcal{B}$ is the Borel $\sigma$-algebra. Additionally, for any tuple $w := (w_i)_{i \in \I}$ indexed by a finite index set $\I$ and any subset $I \subseteq \I$ we write $w_I := (w_i)_{i \in I}$. Similarly, we define $\mathcal{W}_I := \prod_{i \in I} \mathcal{W}_i$ for any collection of spaces $(\mathcal{W}_i)_{i \in \I}$.

\subsection{Definition}

Causal reasoning rests on the knowledge of a \textit{structural causal model} (SCM), which represents the causal relationships between the studied variables.

\begin{definition}
Let $\I$ and $\J$ be two disjoint finite index sets, and write $\mathcal{V} := \prod_{i \in \I} \mathcal{V}_i \subseteq \R^{|\I|}$, $\mathcal{U} := \prod_{i \in \J} \mathcal{U}_i \subseteq \R^{|\J|}$ for two measurable product spaces. A \emph{structural causal model} $\mathcal{M}$ is a couple $\langle U, G \rangle$ where:

\begin{enumerate}
    \item $U : \Omega \to \mathcal{U}$ is a vector of random variables, sometimes called the \emph{random seed};
    \item $G = \{G_i\}_{i \in \I}$ is a collection of measurable $\R$-valued functions, where for every $i \in I$ there exist two subsets of indices $\operatorname{Endo}(i) \subseteq \I$ and $\operatorname{Exo}(i) \subseteq \J$, respectively called the \emph{endogenous} and \emph{exogenous parents} of $i$, such that $G_i : \mathcal{V}_{\operatorname{Endo}(i)} \times \mathcal{U}_{\operatorname{Exo}(i)} \to \mathcal{V}_i$.
\end{enumerate}

A random vector $V : \Omega \to \mathcal{V}$ is a solution of $\mathcal{M}$ if for every $i \in \I$
\begin{equation}\label{eq:causal_eq}
    V_i \stackrel{\P-a.s.}{=} G_i(V_{{\text{Endo}}(i)},U_{\text{Exo}(i)}).
\end{equation}
The collection of equations defined by \eqref{eq:causal_eq} and characterized by $G$ and $U$ are called the \emph{structural equations}. By identifying $G$ to a measurable vector function $G : \mathcal{V} \times \mathcal{U} \to \mathcal{V}$, we compactly write that $V$ is a solution of $\mathcal{M}$ if $V \stackrel{\P-a.s.}{=} G(V,U)$.

\end{definition}

A structural causal model can be seen as a generative model. The variables in $U$ are said to be \textit{exogenous} as they are imposed \emph{a priori} by the model. In contrast, the variables in a solution $V$ are said to be \textit{endogenous} as they
are outputs of the model determined through the structural equations. In practice, the endogenous variables represent observed data, while the exogenous ones model latent background phenomena. Note that compared to \cite{bongers2021foundations}, we do not assume the $(U_j)_{j \in \J}$ to be mutually independent.

\begin{figure}[h!]
    \centering
    \begin{tikzpicture}[-latex ,
    state/.style ={circle ,top color =white,
    draw}]
    \node[state] (V1){$V_1$};
    \node[state] (V2) [above right =of V1] {$V_2$};
    \node[state] (V3) [below right =of V1] {$V_3$};
    \node[state] (U1) [left =of V1] {$U_1$};
    \node[state] (U2) [right =of V2] {$U_2$};
    \node[state] (U3) [right =of V3] {$U_3$};
    \path (V1) edge (V2);
    \path (V1) edge (V3);
    \path (V2) edge (V3);
    \path (U1) edge (V1);
    \path (U2) edge (V2);
    \path (U3) edge (V3);
    \end{tikzpicture}
    \caption{Example of causal graph}
    \label{fig:scm}
\end{figure}
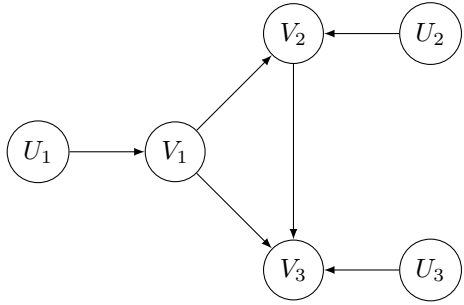

The structural equations specify the causal dependencies between all these variables and are frequently illustrated by the directed graph defined as follows: the set of nodes is $\I \cup \J$, and a directed edge points from node $k$ to node $l$ if and only if $l \in \I$ and $k \in \text{Endo}(l) \cup \text{Exo}(l)$ (we say that $k$ is a parent of $l$). For clarity, we often will substitute the indexes $i \in \I$ or $j \in \J$ for the variables $V_i$ or $U_j$, in particular when drawing such a graph (see Figure~\ref{fig:scm}). Also, similarly to \cite{bongers2021foundations}, we will use in practice non-disjoint subsets $\I$ and $\J$ of duplicated natural integers for the sake of clarity. The example below illustrates the above notations and definitions.

\begin{example}\label{ex:scm}
Consider a simple SCM $\M := \langle U, G \rangle$ where $U := (U_1,U_2,U_3)$ is an arbitrary random vector, and such that $G$ is defined by
\[
    G_1(u_1) := u_1, \quad
    G_2(v_1,u_2) := v_1 + u_2, \quad
    G_3(v_1,v_2,u_3) := v_1 + v_2 + u_3.
\]
Figure~\ref{fig:scm} represents the corresponding graph. By definition, finding a solution $V := (V_1,V_2,V_3)$ to $\M$ amounts to solving,
\[
    V_1 \stackrel{\P-a.s.}{=} U_1, \quad
    V_2 \stackrel{\P-a.s.}{=} V_1 + U_2, \quad
    V_3 \stackrel{\P-a.s.}{=} V_1 + V_2 + U_3.
\]
Then, we readily obtain an almost-surely unique solution given by
\[
    V_1 \stackrel{\P-a.s.}{=} U_1, \quad
    V_2 \stackrel{\P-a.s.}{=} U_1 + U_2, \quad
    V_3 \stackrel{\P-a.s.}{=} 2 U_1 + U_2 + U_3.
\]
\end{example}

According to \citep[Theorem 3.3]{bongers2021foundations}, a model $\M := \langle U, G \rangle$ admits a solution if and only if it is \emph{solvable}, that is there exists a measurable function $\Gamma : \mathcal{U} \to \V$ such that $V \aseq \Gamma(U) \implies V \aseq G(V,U)$. Solvability signifies that a solution $V$ can be expressed solely in terms of $U$, as in the above example. Note that SCMs are not always solvable \citep[Example 2.4]{bongers2021foundations}. For the sake of convenience, we make in the rest of the paper the common assumption that the considered models are \emph{acyclic}, meaning that their graphs do not contain any cycles:
\begin{description}
    \item[(A)\namedlabel{Acyclic}{\textbf{(A)}}] \textit{The structural causal model $\mathcal{M}$ induces a directed \emph{acyclic} graph (DAG).}
\end{description}
Acyclicity entails \emph{unique solvability} of the SCM \cite[Proposition 3.4]{bongers2021foundations}, in the sense that Equation~$\eqref{eq:causal_eq}$ admits a unique solution up to $\P$-negligible sets (as in Example~\ref{ex:scm}). In particular, the generated distribution on the endogenous variables is unique. We will abusively refer to a solution as \emph{the} solution of the SCM.


Essentially, causal structures capture the assumption that features are not independently manipulable. As we detail next, they enable to understand the downstream effect of fixing some variables to certain values onto non-intervened variables.

\subsection{Do-intervention}

The so-called do-calculus embodies mathematically the fundamental distinction between causation and correlation. While standard probability theory can only account for correlations through conditioning, do-calculus allows for \emph{intervening} on variables through the do-operator. Concretely, a do-intervention is an operation mapping any model $\mathcal{M}$ to an alternative one by modifying the generative process.

\begin{definition}\label{def:dointervention}

Let $\mathcal{M} = \langle U, G \rangle$ be an SCM, $I \subseteq \I$ a subset of endogenous variables, and $v_I \in \mathcal{V}_I$ a value. The action $\operatorname{do}(I,v_I)$ defines the modified model $\mathcal{M}_{\operatorname{do}(I,v_I)} = \langle U, \tilde{G} \rangle$ where $\tilde{G}$ is given by

$$
    \tilde{G_i} := \begin{cases}
                    v_i \text{ if } i \in I,\\
                    G_i \text{ if } i \in \I \setminus I.
                   \end{cases}  
$$

\end{definition}

The model surgery described in Definition \ref{def:dointervention} consists in enforcing a state of things by substituting a set of endogenous variables by fixed values while keeping all the rest of the causal mechanism equal. By definition, do-interventions respect the exogeneity of the random seed since $U$ remains unchanged. This  transcribes the causal principle that acting upon endogenous phenomenons does not affect exogenous ones. Provided it is solvable, the modified model $\M_{\operatorname{do}(I,v_I)}$ generates a new distribution of endogenous variables, describing an alternative world where every $V_i$ for $i \in I$ is set to value $v_i$.

Note that do-interventions preserve acyclicity. Therefore, if an SCM $\M$ satisfies \ref{Acyclic}, then $\M_{\operatorname{do}(I,v_I)}$ also satisfies \ref{Acyclic}. Going further, if $V$ is the solution of an acyclic $\M$, we can non-ambiguously define (up to $\P$-negligible sets) its intervened counterpart $V_{\operatorname{do}(I,v_I)}$ solution to $\M_{\operatorname{do}(I,v_I)}$. All in all, \ref{Acyclic} enables to work in a convenient setting where the output of a causal model as well as the ones of its intervened counterparts are always well-defined. This implication enables to clarify the notations: in the sequel we write $\operatorname{do}(V_I = v_I)$ for the operation $\operatorname{do}(I,v_I)$, and use the subscript $V_I=v_I$ to indicate results of this operation. Crucially, intervening does not amount to conditioning in general, that is $\mathcal{L}(V \mid V_I=v_I) \neq \mathcal{L}(V_{V_I=v_I})$.  This means that causal outcomes may not be observable and hence require a known causal model to be inferred, as exemplified below.

\begin{figure}[h!]
    \centering
    \begin{tikzpicture}[-latex ,
    state/.style ={circle ,top color =white,
    draw}]
    \node[state] (V1){$\tilde{V}_1$};
    \node[state] (V2) [above right =of V1] {$\tilde{V}_2$};
    \node[state] (V3) [below right =of V1] {$\tilde{V}_3$};
    \node[state] (U1) [left =of V1] {$U_1$};
    \node[state] (U2) [right =of V2] {$U_2$};
    \node[state] (U3) [right =of V3] {$U_3$};
    \path (V1) edge (V3);
    \path (V2) edge (V3);
    \path (U1) edge (V1);
    \path (U3) edge (V3);
    \end{tikzpicture}
    \caption{Intervened counterpart of Figure~\ref{fig:scm} after $\operatorname{do}(V_2=0)$}
    \label{fig:do_scm}
\end{figure}
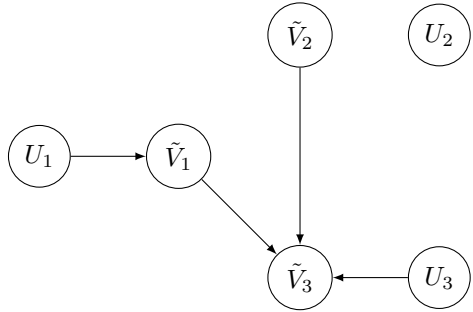

\begin{example}\label{ex:do_scm}
Let $\M := \langle U, G \rangle$ be the SCM from Example~\ref{ex:scm} and consider the do-intervention $\operatorname{do(V_2 = 0)}$. This defines the intervened model $\M_{V_2 = 0} := \langle U, \tilde{G} \rangle$ where
\[
    \tilde{G}_1(u_1) := u_1, \quad
    \tilde{G}_2(v_1,u_2) := 0, \quad
    \tilde{G}_3(v_1,v_2,u_3) := v_1 + v_2 + u_3.
\]
Figure~\ref{fig:do_scm} represents the graph after surgery. The modified structural equations on a solution $\tilde{V} := (\tilde{V}_1,\tilde{V}_2,\tilde{V}_3)$ are
\[
    \tilde{V}_1 \stackrel{\P-a.s.}{=} U_1, \quad
    \tilde{V}_2 \stackrel{\P-a.s.}{=} 0, \quad
    \tilde{V}_3 \stackrel{\P-a.s.}{=} \tilde{V}_1 + \tilde{V}_2 + U_3.
\]
Then, we readily obtain that the almost-surely unique solution is given by
\[
    \tilde{V}_1 \stackrel{\P-a.s.}{=} U_1, \quad
    \tilde{V}_2 \stackrel{\P-a.s.}{=} 0, \quad
    \tilde{V}_3 \stackrel{\P-a.s.}{=} U_1 + U_3.
\]
Assuming that $U_1, U_2, U_3$ are mutually independent we have
$\mathcal{L}(V_1 \mid V_2 = 0) = \mathcal{L}(U_1 \mid U_1 + U_2 = 0) = \mathcal{L}(-U_2)$ while $\mathcal{L}(\tilde{V}_1) = \mathcal{L}(U_1)$. Therefore, $\mathcal{L}(V_1 \mid V_2 = 0) \neq \mathcal{L}(\tilde{V}_1)$ in general.

\end{example}

In Section \ref{sec:models}, we will explain how the do-operator enables counterfactual inference from a causal model. We now turn to the second mathematical theory of interest for our work: mass transportation.

\section{Mass transportation}\label{sec:transport}

We firstly introduce the necessary background on mass (or measure) transportation. Then, we detail the specific case of optimal transport. 

\subsection{Definition}

In probability theory, the problem of mass transportation amounts to constructing a joint distribution namely a \emph{coupling}, between two marginal probability measures. Suppose that each marginal distribution is a sand pile in the ambient space. A coupling is a \emph{mass transportation plan} transforming one pile into the other, by specifying how to move each elementary sand mass from the first distribution so as to recover the second distribution. Alternatively, we can see a coupling as a random matching which pairs start points to end points between the respective supports with a certain weight. Formally, let $P, Q$ be both Borel probability measures on $\R^d$, whose respective supports are denoted by $\text{supp}(P)$ and $\text{supp}(Q)$. We recall that the support is the set of points $x \in \R^d$ such that every open neighbourhood of $x$ has a positive probability. A coupling between $P$ and $Q$ is a probability measure $\pi$ on $\R^d \times \R^d$ admitting $P$ as first marginal and $Q$ as second marginal, precisely  $\pi(A\times\R^d)=P(A)$ and  $\pi(\R^d\times B)=Q(B)$ for all measurable sets $A,B\subseteq \R^d$. Throughout the paper, we denote by $\Pi(P,Q)\subset \mathcal{P}(\R^d\times \R^d)$ the set of joint distributions whose marginals coincide with $P$ and $Q$ respectively.

A coupling $\pi \in \Pi(P,Q)$ is said to be \emph{deterministic} if each instance from the first marginal is paired with probability one to an instance of the second marginal. Such a coupling can be identified with a measurable map $T : \R^d \to \R^d$ that \emph{pushes forward} $P$ to $Q$, that is $Q(B) := P (T^{-1}(B))$ for any measurable set $B\subseteq \R^d$. This property, denoted by $T_\sharp P = Q$, means that if the law of a random variable $Z$ is $P$, then the law of $T(Z)$ is $Q$. To make the relation with random couplings, we also introduce the action of couples of functions on probability measures. For any pairs of functions $T_1, T_2 : \R^d \to \R^d$, we define $(T_1 \times T_2) : \R^d \to \R^d \times \R^d, x \mapsto (T_1(x),T_2(x))$. As such, $(T_1 \times T_2)_\sharp P$ denotes the law of $\big(T_1(Z),T_2(Z)\big)$ where $\mathcal{L}(Z) = P$. This coupling admits ${T_1}_\sharp P$ and ${T_2}_\sharp P$ as first and second marginal respectively. Thus, the deterministic coupling $\pi$ between $P$ and $Q$ characterized by a push-forward operator $T$ satisfying $T_\sharp P = Q$ can be written as $\pi = (I \times T)_\sharp P$ where $I$ is the identity function on $\R^d$. This coupling matches a given instance $x \in \text{supp}(P)$ to $T(x) \in \text{supp}(Q)$ with probability 1.

\subsection{Optimal transport}\label{sec:OT}

We recall here some basic knowledge on optimal transport theory, which is the mass transportation approach we focus on in this work, and refer to \citep{villani2003topics,villani2008optimal} for further details. Optimal transport restricts the set of feasible couplings between two marginals by isolating ones that are optimal in some sense.

\subsubsection{Arbitrary cost}

The \emph{Monge formulation} of the optimal transport problem with general cost $c : \R^d \times \R^d \to \R$ is the optimization problem
\begin{equation}\label{eq:monge}
    \min_{T :\ T_\sharp P = Q} \int_{\R^d} c(x,T(x)) \mathrm{d} P(x).
\end{equation}
We refer to solutions to \eqref{eq:monge} as \emph{optimal transport maps} between $P$ and $Q$ with respect to $c$; they minimize the effort, quantified by $c$, of moving every elementary mass from $P$ to $Q$. One mathematical complication is that the push-forward constraint renders the problem unfeasible in many general settings, in particular when $P$ and $Q$ are not absolutely continuous with respect to the Lebesgue measure or have unbalanced numbers of atoms.

This issue motivates the following \emph{Kantorovich relaxation} of the optimal transport problem with cost $c$,
\begin{equation}\label{eq:kanto}
    \min_{\pi\in \Pi(P,Q)} \int_{\R^d\times \R^d} c(x,x') \mathrm{d}\pi(x,x').
\end{equation}
Solutions to \eqref{eq:kanto} are \emph{optimal transport plans} (possibly non deterministic) between $P$ and $Q$ with respect to $c$. In contrasts to optimal transport maps, they exist under very mild assumptions, like the non negativeness of the cost. Notice that, since a push-forward operator can be identified to a coupling, the set of feasible solutions to \eqref{eq:monge} is included in the set of feasible solutions to \eqref{eq:kanto}.

\subsubsection{Quadratic cost}\label{sec:brenier}

Optimal transport enjoys a well-established theory, in particular when the ground cost is the squared Euclidean distance $c(x,x'):=\norm{x-x'}^2$ on $\R^d \times \R^d$. Suppose that $P$ is absolutely continuous with respect to the Lebesgue measure in $\R^d$, and that both $P$ and $Q$  have finite second order moments. Theorem~2.12 in \cite{villani2003topics}, originally proved by \cite{cuesta1989notes} and then \cite{brenier1991polar}, states that there exists a unique solution to Kantorovich's formulation of optimal transport \eqref{eq:kanto}, whose form is $(I \times T )_\sharp P$ where $T : \R^d \to \R^d$ solves the corresponding squared Monge problem,

\begin{equation}\label{eq:square_monge}
    \min_{T:\ T_\sharp P = Q} \int_{\R^d} \norm{x-T(x)}^2 \mathrm{d}P(x).
\end{equation}

Although it may not be unique, this optimal transport map $T$ is uniquely determined $P$-almost everywhere, and we will abusively refer to it as \emph{the} solution to Problem~\eqref{eq:square_monge}. Crucially, this map coincides $P$-almost everywhere with the gradient of a convex function. Moreover, according to \cite{mccann1995}, under the sole assumption that $P$ is absolutely continuous with respect to the Lebesgue measure, there exists only one (up to $P$-negligible sets) gradient of a convex function $\nabla \phi$ satisfying the push-forward condition $\nabla \phi_\sharp P = Q$. We combine Brenier's and McCann's theorems into the following lemma, which simplifies the search for the solutions to \eqref{eq:square_monge}.

\begin{lemma}\label{lm:connect}
Assume that $P$ is absolutely continuous with respect to the Lebesgue measure, and that both $P$ and $Q$ have finite second order moments. Then, a measurable map $T : \operatorname{supp}(P) \to \operatorname{supp}(Q)$ is a solution to \eqref{eq:square_monge} if and only if it satisfies the two following conditions:
\begin{enumerate}
    \item $T_\sharp P = Q$,
    \item there exists a convex function $\phi : \R^d \to \R$ such that $T = \nabla \phi$ $P$-almost everywhere.
\end{enumerate}
\end{lemma}
This result will play a key role in Section \ref{sec:coincide} to prove a link between optimal transport and causality.

\subsubsection{Implementation}\label{sec:estimation}

In practice, we do not know the measures $P$ and $Q$ but have access to empirical observations. This naturally raises the questions of building relevant data-driven approximations, or estimators, of the optimal transport plans, and of what should be required to ensure statistical guarantees. In this section, we briefly present the computational aspects of optimal transport, and refer to \citep{peyre2019computational} for a complete overview.

Concretely, consider two samples of i.i.d. observations $\{x_1,\ldots,x_n\}$ and $\{x'_1,\ldots,x'_m\}$ drawn from respectively $P$ and $Q$. These samples define the empirical measures $P^n = n^{-1} \sum^n_{i=1} \delta_{x_i}$ and $Q^m = m^{-1} \sum^m_{i=1} \delta_{x'_i}$, where $\delta_x$ denotes the Dirac measure at point $x$. Then, the standard way to estimate an optimal transport plan between the marginals $P$ and $Q$ is to solve the Kantorovich formulation \eqref{eq:kanto} between their empirical counterparts $P^n$ and $Q^m$. By identifying a discrete coupling to a matrix, we write this problem as,
\begin{equation}\label{eq:empirical_kanto}
    \min_{\pi \in \Sigma(n,m)} \sum^n_{i=1} \sum^m_{j=1} c(x_i,x'_j) \pi(i,j),
\end{equation}
where $\Sigma(n,m) := \{ \pi \in {\R}^{n \times m}_+ \mid \sum^m_{j=1} \pi(i,j) = n^{-1} \text{ and } \sum^n_{i=1} \pi(i,j) = m^{-1}\}$. Note that empirical transport plans are statistically consistent. This means that if the Kantorovich problem \eqref{eq:kanto} admits a unique solution $\pi$, then a sequence $\{\pi^{n,m}\}_{n,m \in \N}$ of solutions to Problem~\eqref{eq:kanto} converges weakly to $\pi$ as $n$ and $m$ increase to infinity \citep[Theorem 5.19]{villani2008optimal}. This property is crucial to ensure statistical guarantees in optimal-transport frameworks. We emphasize that even if a solution to Problem~\eqref{eq:empirical_kanto} is necessarily non-deterministic as soon as $n \neq m$, the corresponding solution to Problem~\eqref{eq:kanto} can be deterministic.

The main challenge when working with empirical optimal-transport solutions is that they are expensive in both computational complexity and memory: solving \eqref{eq:empirical_kanto} typically requires $\mathcal{O}((n+m)n m\log(n+m))$ computer operations, and the solution is stored as an $n \times m$ matrix, which can limit the application on large datasets. Our implementation (see the experiments in Section~\ref{sec:numerical}) exploits the sparsity of the transport matrix to avoid overloading the memory and to speed-up the evaluation of optimal-transport-based metrics. One could also consider entropic regularization schemes to accelerate the computation of a solution to reach $\mathcal{O}(n m)$ operations \citep{cuturi2013sinkhorn}. However, the obtained approximation of the transport matrix is typically non sparse, hence contains many non-zero coefficients, which precludes memory-efficient implementations. This is why we address only standard optimal transport in our numerical experiments.


\section{Counterfactual models}\label{sec:models}

We now have all the tools to focus on the main subject of this paper: counterfactual reasoning. As mentioned in the introduction, both causality and transport techniques have been used for this purpose. However, a yet non-appreciated aspect is that these frameworks can be written in a common formalism; this is what this section addresses. More precisely, we propose the definition of \emph{counterfactual models}, mathematical objects encoding the probabilities of all counterfactual statements with respect to modifications of one variable, and detail how to construct them with respectively causal models and mass-transportation methods.

\subsection{Problem setup}\label{sec:setup}

Set $d \geq 1$, and define the random vector $V := (X,S) \in \R^{d+1}$, where the variables $X : \Omega \to \mathcal{X} \subseteq \R^d$ represent some observed features, while the variable $S: \Omega \to \mathcal{S} \subset \R$ can be subjected to interventions. For simplicity, we assume that $\S$ is finite such that for every $s \in \S$, $\P(S=s)>0$. We consider the problem of computing the potential outcomes of $X$ when changing $S$. Typically, $S$ represents the sensitive, protected attribute in fairness settings, or the treatment status in the potential-outcome framework. Suppose for instance that the event $\{X=x,S=s\}$ is observed, and set $s' \neq s$. We aim at answering the counterfactual question: \emph{had $S$ been equal to $s'$ instead of $s$, what would have been the value of $X$?} Critically, because of correlations or structural relations between the variables, computing the alternative state does not amount to change the value of $S$ while keeping the features $X$ equal.

\subsection{Structural counterfactuals}\label{sec:structural}

Answering the counterfactual question from Section \ref{sec:setup} with Pearl's framework requires to assume causal dependencies between $X$ and $S$. Formally, suppose that $V = (X,S) \in \R^{d+1}$ is the unique solution to an SCM $\M = \langle U, G \rangle$ satisfying the acyclicity assumption $\ref{Acyclic}$. We recall that each \emph{endogenous} variable $V_k$ is then defined (up to sets of probability zero) by the structural equation
\begin{equation*}
    V_k \stackrel{\P-a.s.}{=} G_k\left(V_{\text{Endo}(k)},U_{\text{Exo}(k)}\right),
\end{equation*}
where $G_k$ is a real-valued measurable function, $U$ is a vector of \emph{exogenous} variables, while $V_{\text{Endo}(k)}$ and $U_{\text{Exo}(k)}$ denote respectively the endogenous and exogenous parents of $V_k$. In the following, we denote by $U_X$ and $U_S$ the exogenous parents of respectively $X$ and $S$. We write $X_{S=s}$ the intervened counterpart of $X$ through the do-intervention $\operatorname{do}(S=s)$, that is after replacing the structural equation on $S$ by $S=s$ while keeping the rest of the causal mechanism equal.

Then, we introduce the following  notations to formalize the contrast between interventional, counterfactual and factual outcomes. For $s,s' \in \S$ we define three probability distributions. Firstly, $\mu_s := \mathcal{L}(X \mid S=s)$ is the distribution of the \emph{factual} $s$-instances. This observable measure describes the possible values of $X$ such that $S=s$, and we write $\X_s$ for its support. Secondly, we denote by $\mu_{S=s} := \mathcal{L}(X_{S=s})$ the distribution of the \emph{interventional} $s$-instances. It describes the alternative values of $X$ in a world where $S$ is forced to take the value $s$. On the contrary to the factual distribution, the interventional distribution is in general not observational, in the sense that we cannot draw empirical observations from it. Finally, we define by $\mu_{\langle s'|s \rangle} := \mathcal{L}(X_{S=s'} \mid S=s)$ the distribution of the \emph{counterfactual} $s'$-instances given $s$. It describes what would have been the factual instances of $\mu_s$ \emph{had $S$ been equal to $s'$ instead of $s$}. According to the \emph{consistency rule} \citep{pearl2016causal}, the factual and counterfactual distributions coincide when $s=s'$, that is $\mu_s = \mu_{\langle s|s \rangle}$. However, when $s \neq s'$, the counterfactual distribution $\mu_{\langle s'|s \rangle}$ is generally not observable.

\subsubsection{Definition}

Using the above notation, our problem can be framed as: having observed an $x \in \X_s$, determining the probability of the counterfactual outcome $x' \in \text{supp}(\mu_{\langle s'|s \rangle})$. Pearl originally answered this question with the following \emph{three-step procedure}: (1) set a prior $\mathcal{L}(U)$ for the SCM, (2) compute the posterior distribution $\mathcal{L}(U \mid X=x,S=s)$, and (3) solve the structural equations after the intervention $\operatorname{do}(S=s')$ with $\mathcal{L}(U \mid X=x,S=s)$ as input. This leads to the following formal definition of \emph{structural counterfactuals}, adapted from \citep[Chapter 4]{pearl2016causal}.

\begin{definition}\label{def:3steps}

Let $\M$ satisfy \ref{Acyclic}. For an observed evidence $\left\{X=x,S=s\right\}$ and an intervention $\operatorname{do}(S=s')$, the \emph{structural counterfactuals} of $X$ are characterized by the probability distribution $\mu_{\langle s'|s \rangle}(\cdot|x)$ defined as

$$
    \mu_{\langle s'|s \rangle}(\cdot|x) := \mathcal{L}(X_{S=s'} \mid X=x,S=s).
$$

\end{definition}

In general, the structural counterfactuals of a single instance are not necessarily \emph{deterministic}, that is characterized by a degenerate distribution, but belong to a set of possible outcomes with probability weights. This comes from the fact that several values of $U$ can generate a same observation $\{X=x,S=s\}$. This means that, according to Pearl's causal reasoning, there is not necessarily a one-to-one correspondence between factual instances and counterfactual counterparts, but a collection of weighted correspondences described by the distribution of structural counterfactuals.

\subsubsection{Mass-transportation viewpoint}

While the mainstream literature on causality generally operates with the definition of structural counterfactuals given by the three-strep procedure \citep{kusner2017counterfactual,barocas-hardt-narayanan}, we focus in this paper on a \emph{mass-transportation viewpoint} of counterfactuals, formalized by the following definition.

\begin{definition}\label{def:ctp} Let $\M$ satisfy \ref{Acyclic}. For every $s,s' \in \S$, the \emph{structural counterfactual coupling} between $\mu_s$ and $\mu_{\langle s'|s \rangle}$ is given by
$$
\pi^*_{\langle s'|s \rangle} := \mathcal{L}\big((X, X_{S=s'}) \mid S=s\big).
$$
\noindent
We call the collection of couplings $\Pi^* := \{\pi^*_{\langle s'|s \rangle}\}_{s,s'\in \S}$ the \emph{structural counterfactual model} on $X$ with respect to $S$.
\end{definition}
In this formalism, the quantity $\mathrm{d} \pi_{\langle s' | s\rangle}(x,x')$ is the elementary probability of the counterfactual statement \emph{had $S$ been equal to $s'$ instead of $s$ then $X$ would have been equal to $x'$ instead of $x$}. As such, a counterfactual model characterizes the distribution of all the cross-world statements on $X$ with respect to changes of $S$. Note that each realization of $\pi^*_{\langle s'|s \rangle}$, that is each pair of factual instance and counterfactual counterpart, is generated by a same possible value of $\mathcal{L}(U_X \mid S=s)$.

We point out that Definitions~\ref{def:3steps} and \ref{def:ctp} characterize the exact same counterfactual statements, the formal link being $\mathrm{d}\pi^*_{\langle s'|s \rangle}(x,x') = \mu_{\langle s'|s \rangle}(x'|x)  \mathrm{d} \mu_s(x)$. In particular, there is an equivalence between $\mu_{\langle s'|s \rangle}(\cdot|x)$ narrowing down to a single value for every $x \in \X_s$ and $\pi^*_{\langle s'|s \rangle}$ being a deterministic coupling. Assumptions rendering single-valued counterfactuals will be studied in Section \ref{sec:deterministic}. We also note that this joint-probability-distribution perspective of Pearl's counterfactuals concurs with the one from \cite[Section 2.5]{bongers2021foundations}.




\subsection{Transport-based counterfactuals}\label{sec:surrogate}

The main issue of structural counterfactuals, which will be widely discussed in Section~\ref{sec:discussion}, comes from the causal model being unknown in practice. Thus, the necessity to make counterfactual frameworks feasible naturally raises the question of finding good surrogates to causal counterfactuals. We have seen that the problem of assessing counterfactual statements about $X$ with respect to interventions on $S$ using causal models could be reduced to knowing a collection of random mappings from factual distributions $\{\mu_s\}_{s \in \S}$ towards counterfactual distributions $\{\mu_{\langle s'|s \rangle}\}_{s,s' \in \S}$. This perspective suggests that mass-transportation techniques can be natural substitutes for structural counterfactual reasoning, as they remedy to the aforementioned issues.


\subsubsection{Definition}

In \citep{black2020fliptest}, the authors mimicked the structural account of counterfactuals by computing alternative instances using a deterministic optimal transport map. Extending their idea, we propose a more general framework where the counterfactual operation switching $S$ from $s$ to $s'$ can be seen as a mass transportation plan, not necessarily optimal-transport based and not necessarily deterministic, between two distributions.\footnote{In \citep[Section 7.2]{asher2022counterfactual}, we present this view of counterfactuals from a logic perspective.} In the following, $t : \R^d \times \R^d \to \R^d \times \R^d, (x,x') \mapsto (x',x)$ denotes the permutation function.

\begin{definition}\label{def:TCFM}
A \emph{transport-based counterfactual model} is a collection of couplings $\Pi := \left\{\pi_{ \langle s'|s \rangle}\right\}_{s,s' \in \S}$ satisfying for every $s,s' \in \S$,
\begin{itemize}
    \item[(i)] $\pi_{ \langle s'|s \rangle} \in \Pi(\mu_s,\mu_{s'})$;
    \item[(ii)] $\pi_{ \langle s|s \rangle} = (I \times I)_\sharp \mu_s$;
    \item[(iii)] $\pi_{ \langle s|s' \rangle} = t_\sharp \pi_{ \langle s'|s \rangle}$.
\end{itemize}
An element of $\Pi$ is called a \emph{counterfactual coupling}. We say that $\Pi$ is a \emph{random} counterfactual model if at least one coupling for $s \neq s'$ is not deterministic. Otherwise, we say that $\Pi$ is a \emph{deterministic} counterfactual model. In the deterministic case, $\Pi$ can be identified almost everywhere to a collection $\T := \left\{T_{\mathsmaller{ \langle s'|s \rangle}}\right\}_{s,s' \in \S}$ of measurable mappings from $\X$ to $\X$ satisfying for every $s,s' \in \S$,
\begin{itemize}
    \item[(i)] ${T_{\mathsmaller{ \langle s'|s \rangle}}}_\sharp \mu_s = \mu_{s'}$;
    \item[(ii)] $T_{\mathsmaller{ \langle s|s \rangle}} = \text{I}$;
    \item[(iii)] $T_{\mathsmaller{ \langle s'|s \rangle}}$ is invertible $\mu_s$-almost everywhere such that $T_{\mathsmaller{ \langle s|s' \rangle}} = T^{-1}_{\mathsmaller{ \langle s'|s \rangle}}$.
\end{itemize}
An element of $\T$ is called a \emph{counterfactual operator}.
\end{definition}

Similarly to structural counterfactual models, these models assign a probability to all the cross-world statements on $X$ with respect to interventions on $S$. By convention, we use the superscript $^*$ to denote \emph{structural} counterfactual models, and no superscript for \emph{transport-based} counterfactual models. The marginal constraint $(i)$ in Definition~\ref{def:TCFM} translates the intuition that a realistic counterfactual operation on $S$ should morph the non-intervened variables $X$ so that their values fit the targeted distribution. In this sense, transport-based models preserve the principle that features are not independently manipulable, but without using causal relations. The symmetry constraints $(ii)$ and $(iii)$ cover the reciprocity intuition we have on counterfactuals counterparts. Remark that in the case of discrete measures, the operation $t_\sharp$ in condition $(iii)$ simply amounts to transposing the associated coupling matrices. Lastly, note that this definition replaces the unobservable, SCM-dependent distributions $\{\mu_{\langle s'|s \rangle}\}_{s,s' \in \S}$ of structural counterfactual models by the observational $\{\mu_{s'}\}_{s' \in \S}$ for feasibility reasons. In Section~\ref{sec:exogenous}, we will see that this approximation makes sense in typical fairness settings where $\mu_{\langle s'|s \rangle} = \mu_{s'}$ for every $s,s' \in \S$.

The adjective \emph{deterministic} refers to the fact that the model assigns to each factual instance a unique counterfactual counterpart. Formally, the counterfactual counterpart of some observation $x \in \X_s$ after changing $S$ from $s$ to $s'$ is given by $x' = T_{\mathsmaller{\langle s' | s \rangle}}(x) \in \X_{s'}$. In contrast, a \emph{random} model allows possibly several counterparts with probability weights. The first interest of considering random couplings is purely conceptual; rendering non necessarily unique the counterfactual counterparts of a single instance has philosophical implications \citep[Section 6.3]{asher2022counterfactual}. Besides, it is consistent with the causal approach which also authorizes non-deterministic counterfactuals. The second---and most critical benefit---is practical. While there always exist random couplings between two distributions, deterministic push-forward mappings (even causally-induced ones) may not exist when the marginals do not have densities, making this relaxation crucial for dealing with non-continuous variables. This makes the extension to random couplings necessary to tackle concrete machine-learning tasks, involving both continuous and discrete covariates. Notably, we rely on random couplings in the numerical experiments from Section~\ref{sec:application}.

\subsubsection{Choosing a model}

One challenge for the transport-based approach is to choose the model appropriately in order to define a relevant notion of counterpart. There possibly exists an infinite number of admissible counterfactual models in the sense of Definition \ref{def:TCFM}, many of them being inappropriate. As an illustration, consider the family of trivial couplings, namely $\{\mu_s \otimes \mu_{s'}\}_{s,s'\in \S}$ where $\otimes$ denotes the factorization of measures. Though it is a well-defined transport-based counterfactual model, it is not intuitively justifiable as it completely decorrelates factual and counterfactual instances. To sum-up, a transport-based counterfactual model must be both \emph{intuitively justifiable} and \emph{computationally feasible}.

We argue that optimal-transport solutions are tailored couplings with respect to both criteria. Optimal transport has become the most popular tool in statistics-related fields to define couplings between distributions when no canonical choice is available, as in generative modeling \citep{balaji2020robust}, domain adaptation \citep{courty2014domain,courty2017joint,redko2019optimal,rakotoarison2022learning}, and transfer learning \citep{gayraud2017optimal,peterson2021transfer} thanks to significant advances in computational schemes. Additionally, as argued by \cite{black2020fliptest}, generating a counterfactual operation by solving the optimal-transport Problem \eqref{eq:monge} leads to intuitively relevant counterfactuals, as they are obtained by minimizing a metric between paired instances (transcribing the Lewisian most-similar-alternative-world principle) while preserving the probability distributions (ensuring distributional faithfulness). Moreover, deterministic optimal transport for the quadratic cost (see Section~\ref{sec:brenier}) has remarkable properties. According to Lemma~\ref{lm:connect}, solutions to Problem~\eqref{eq:square_monge} are gradients of convex functions, which extends the notion of non-decreasing function to several dimensions. In particular, the optimal transport map in dimension one is the quantile-preservation map between univariate distributions. This behaviour has notably inspired constructions of multivariate notions of quantile based on optimal transport \citep{chernozhukov2017monge,hallin2021distribution,ghosal2022multivariate}. It also makes sense in counterfactual reasoning where, without further information on the data-generation process, preserving the quantile from on marginal to the other is an intuitive definition of the counterfactual counterpart.  For the sake of illustration, Section~\ref{sec:examples} below provides several examples of optimal transport applied to counterfactual reasoning.

In Section~\ref{sec:coincide} we will further justify the pertinence of \emph{optimal-transport-based} counterfactual models by showing that they coincide with structural counterfactual models under some assumptions. However, the scope of Definition~\ref{def:TCFM} goes beyond solutions to standard optimal-transport problems, allowing other transport methods and as such more possible counterfactual models. The purpose of this generalization is partly theoretic: in the future, one could propose an original matching technique and justify its relevance compared to optimal transport. In particular, the couplings mentioned in \citep[Chapter 1]{villani2008optimal} as well as diffeomorphic registration mappings \citep{joshi2000landmark,beg2005computing} are possible candidates we do not investigate in this paper. Additionally, this generalization permits the use of regularized optimal transport \citep{cuturi2013sinkhorn}, which deviates from the original formulation of Problem~\eqref{eq:kanto}, to accelerate computations. Note in passing that solutions to regularized optimal transport, which are non deterministic, define adequate transport-based counterfactual models thanks to Definition~\ref{def:TCFM} taking into account random couplings. Lastly, we will see in Section~\ref{sec:exogenous} that structural counterfactual models are transport-based counterfactual models---but not necessarily optimal-transport-based---under some assumptions. 

\subsection{Examples}\label{sec:examples}

Now that we gave definitions and insights on counterfactual models,  let us study two concrete examples on real data.

\subsubsection{Law dataset}\label{sec:law}

We start by focusing on the Law School Admission Council dataset which gathers statistics from 163 US law schools and more than 20,000 students, including four variables: the race $S$, the entrance-exam score $X_1$, the grade-point average before law school $X_2$, and the first-year average grade $Y$. The end goal is to predict the first-year grade $Y$ from the other features $(X,S)$. Similarly to \cite{russell2017when}, we consider a fairness setting where the race plays the role of a protected, sensitive attribute which should not be discriminated against, and we restrict to only black ($S=0$) and white ($S=1$) students. Counterfactual reasoning has become popular in such algorithmic fairness tasks to either ensure or test that, for example, had a black student been white, the output would have been the same. This requires a model to compute the counterfactual counterparts of any students after changing their skin colors.

\begin{figure}[h]
     \centering
     \begin{subfigure}[b]{0.47\textwidth}
         \centering
         \includegraphics[width=\textwidth]{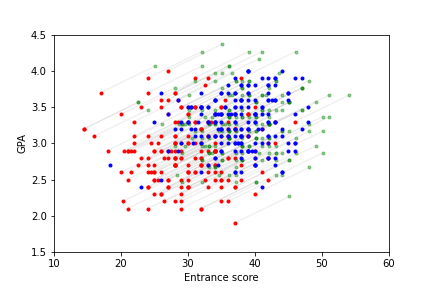}
         \caption{$T^*_{\langle 1|0 \rangle}$}
         \label{fig:law_causal_0to1}
     \end{subfigure}
     \hfill
     \begin{subfigure}[b]{0.47\textwidth}
         \centering
         \includegraphics[width=\textwidth]{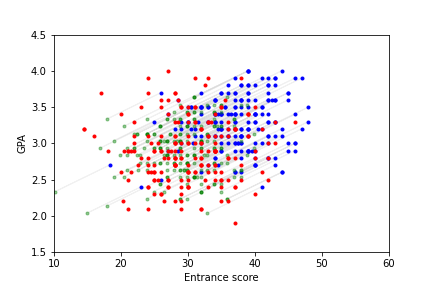}
         \caption{$T^*_{\langle 0|1 \rangle}$}
         \label{fig:law_causal_1to0}
     \end{subfigure}
     
          \centering
     \begin{subfigure}[b]{0.47\textwidth}
         \centering
         \includegraphics[width=\textwidth]{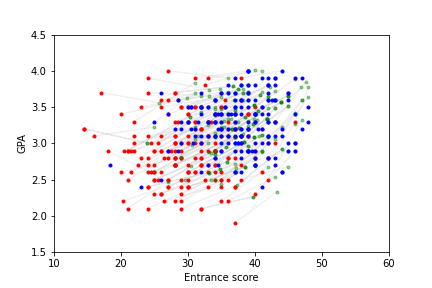}
         \caption{$T_{\langle 1|0 \rangle}$}
         \label{fig:law_OT_0to1}
     \end{subfigure}
     \hfill
     \begin{subfigure}[b]{0.47\textwidth}
         \centering
         \includegraphics[width=\textwidth]{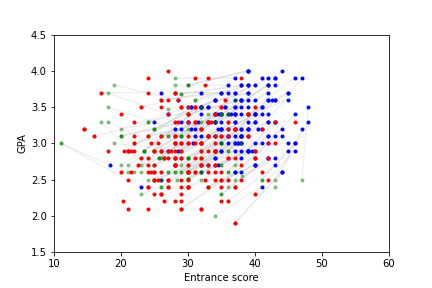}
         \caption{$T_{\langle 0|1 \rangle}$}
         \label{fig:law_OT_1to0}
     \end{subfigure}

    \caption{Counterfactual models for the Law dataset. The red sample represents 200 factual black students while the blue sample represents 200 factual white students. The green sample depicts counterfactual instances: the first column (Figures~\ref{fig:law_causal_0to1} and \ref{fig:law_OT_0to1}) has white counterfactual students; the second column (Figures~\ref{fig:law_causal_1to0} and \ref{fig:law_OT_1to0}) has black counterfactual students. The lightgray lines describe the coupling between factual and counterfactual instances.}
    \label{fig:models}
\end{figure}

First, we consider a structural counterfactual model. This requires a causal model: \cite{russell2017when} proposed the following SCM for the dataset,
\[
\begin{cases}
X_1 = b_1 + w_1 S + U_1,\\
X_2 = b_2 + w_2 S + U_2,\\
S = U_S,\\
U_S, U_1, U_2~\text{independent},
\end{cases}
\]
where $b := (b_1,b_2)$ and  $w := (w_1,w_2)$ are deterministic $\R^2$ parameters obtained by adjusting linear-regression models component-wise. Let us now calculate the induced structural counterfactual model by applying Definition~\ref{def:ctp}. The coupling from $S=0$ to $S=1$ is given by
\[
\pi^*_{\langle 1|0 \rangle} := \mathcal{L}\big((X, X_{S=1}) \mid S=0\big) = \mathcal{L}\big((b+U_X, b+w+U_X)\big) = \mathcal{L}\big((X, X+w) \mid S=0\big).
\]
Conversely, the structural counterfactual coupling from $S=1$ to $S=0$ is
\[
\pi^*_{\langle 0|1 \rangle} := \mathcal{L}\big((X, X_{S=0}) \mid S=1\big) = \mathcal{L}\big((b+w+U_X, b+U_X)\big) = \mathcal{L}\big((X, X-w) \mid S=1\big).
\]
Figures~\ref{fig:law_causal_0to1} and \ref{fig:law_causal_1to0} illustrate the computation of the corresponding counterfactual counterparts on samples. We make two important remarks.

Firstly, generating counterfactual quantities in this case amounts to translating instances of $\mu_0$ by the constant $w$ or conversely translating instances of $\mu_1$ by the constant $-w$. Notably, the two couplings are deterministic: $\pi^*_{\langle 1|0 \rangle}$ and $\pi^*_{\langle 0|1 \rangle}$ are respectively characterized by the mappings $T^*_{\langle 1|0 \rangle}(x) := x + w$ and $T^*_{\langle 0|1 \rangle}(x) := x - w$. Note that there is consequently no need to specify the law of the exogenous variables to compute counterfactual quantities. Section~\ref{sec:deterministic} provides a general analysis of such deterministic settings.

Secondly, the causal model implies that $S \independent U_X$. This critically entails that the counterfactual distributions are observable, since $\mu_{\langle 1 | 0 \rangle} = \mathcal{L}(X_{S=1} \mid S=0) = \mathcal{L}(b+w+U_X \mid S=0) = \mathcal{L}(b+w+U_X \mid S=1) = \mu_1$ and $\mu_{\langle 0 | 1 \rangle} = \mu_0$  analogously. Therefore, the structural counterfactual couplings $\pi^*_{\langle 1|0 \rangle}$ and $\pi^*_{\langle 0|1 \rangle}$ belong respectively to $\Pi(\mu_0,\mu_1)$ and $\Pi(\mu_1,\mu_0)$. Additionally, they are transposed from one another, that is $t_\sharp \pi^*_{\langle 1|0 \rangle} = \pi^*_{\langle 0|1 \rangle}$. This means that the structural counterfactual model $\Pi^* := \{ \pi^*_{\langle 1|0 \rangle}, \pi^*_{\langle 0|1 \rangle}\}$ is a transport-based counterfactual model. Mathematical justifications of these properties will be studied in Section~\ref{sec:exogenous}.

In a second time, we turn to an optimal-transport-based counterfactual model. More precisely, we learn the optimal transport map for the quadratic cost, denoted by $T_{\langle 1|0 \rangle}$, from the black distribution $\mu_0$ towards the white distribution $\mu_1$. In practice, we rely on the Python Optimal Transport (POT) library to compute an approximation of the mapping from data \citep{flamary2021pot}. Note that solving the empirical optimal-transport problem \eqref{eq:empirical_kanto} between samples provides a matching that cannot generalize to new, out-of-sample observations. This is why we employ POT's in-built non-regularized barycentric extension of the empirical solution to obtain a mapping defined everywhere. We use 800 points from each distribution to compute the estimator of $T_{\langle 1|0 \rangle}$ illustrated in Figure~\ref{fig:law_causal_0to1}. The converse counterfactual operation $T_{\langle 0|1 \rangle}$ represented in Figure~\ref{fig:law_OT_1to0} is produced by inversion.

We emphasize that all the couplings in Figure~\ref{fig:models}, be they causal-based or optimal-transport-based, are imperfect approximations, but for different reasons. More precisely, we assumed that a linear causal model generated the data in order to compute the structural counterfactual couplings. However, this model-class assumption is not a perfect fit: in particular, some of the produced counterfactual instances are not realistic, yielding GPA scores exceeding the upper limit of 4.0 points; more generally, while both couplings should have $\mu_0$ and $\mu_1$ for marginals, several counterfactual counterparts do not conform to these distributions. Besides, the translation vector $w$ used in practice is an estimation from data, thereby an approximation of the best linear model fitting the data. The implemented optimal-transport mappings are also mere estimators of the \say{true} mappings between the continuous distributions. Figure~\ref{fig:law_OT_0to1} notably shows poor counterfactual associations for outliers of the red sample, likely due to weak estimation in low-density domains. Nevertheless, the marginal constraint of optimal transport ensures that the generated counterfactuals faithfully fit the data and are therefore plausible. Finally, despite these approximation artifacts, we remark that the causal and optimal-transport couplings have fairly similar behaviours, siding with the observations of \cite{black2020fliptest}. This proximity will be theoretically grounded in Section~\ref{sec:coincide}.

\subsubsection{Body-measurement dataset}\label{sec:body}

We now further illustrate the properties of optimal-transport counterfactuals on a dataset of body measurements from $n_0 = 260$ women and $n_1 = 247$ men. The features of interest are the weight $X_1$ and the height $X_2$, while $S$ encodes the gender. Suppose now that Bob is a 80kg and 190cm man. What would have been Bob's height and weight had he been a woman? Since we do not know the structural relationships between $X$, $S$ and possibly hidden sources of randomness $U$, we follow \cite{black2020fliptest} and rely on mass-transportation techniques to answer this counterfactual question. We proceed as before to estimate the optimal transport map from the male distribution $\mu_1$ towards the female distribution $\mu_0$. Applying this operator to Bob, we obtain that, had he been a women, she would have been 59kg and 177cm.

\begin{figure}[h]
     \centering
     \begin{subfigure}[b]{0.47\textwidth}
         \centering
         \includegraphics[width=\textwidth]{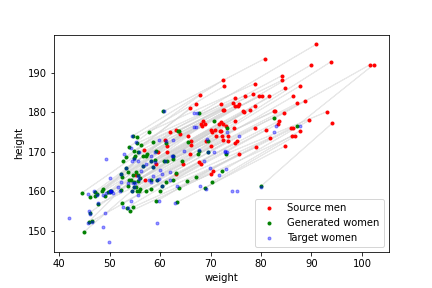}
    \caption{2D optimal-transport matching.}
    \label{fig:ot_map}
     \end{subfigure}
     \hfill
     \begin{subfigure}[b]{0.47\textwidth}
         \centering
         \includegraphics[width=\textwidth]{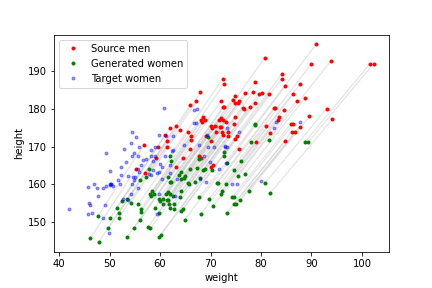}
    \caption{2D component-wise quantile-preservation.}
    \label{fig:quantile}
     \end{subfigure}

    \caption{Body dataset. The red dots represent a data sample of men, while the blue dots represent a data sample of women. The green dots are the estimated counterfactual counterparts of the male sample.}
    \label{fig:body}
\end{figure}

Though it does not have a canonical definition when $d=2$, optimal transport seems visually to preserve the \say{position} of the paired points from one marginal to another. This is due to the optimal map being the unique gradient of a convex function between distributions as previously explained. We underscore in Figure~\ref{fig:body} that optimal transport does not amount to feature-wise quantile-preservation, making it a relevant extension of the notion of order to higher dimension. Notably, preserving the quantile along each coordinate does not satisfy the marginal constraint, yielding counterfactual women not representative of their gender's distribution. 

\subsection{Discussion}\label{sec:discussion}

Counterfactuals have valuable applications in fairness and explainability. One could for example try to learn predictor $h$ designed to make $h(x,0)$ as close as possible to $h(x',1)$ for every counterfactual pair $(x,x')$. This is what \cite{russell2017when} proposed using causal models, and what we implement in Section~\ref{sec:application} using transport-based models. Or, one could test whether a trained predictor $h$ is unfair by checking if $h(x,0)=h(x',1)$ for every counterfactual pair $(x,x')$, which is essentially the procedure of \cite{black2020fliptest} leveraging optimal transport maps. However, the application of counterfactual models raises several issues. We conclude Section~\ref{sec:models} by discussing important drawbacks of the causal account to counterfactual reasoning as well as the limitations of the transport approach.

\subsubsection{Shortcomings of the causal approach}

The main limitation, as for any causal-based framework, is its feasibility. Assuming a known causal model, in particular a fully-specified causal model, is a too strong assumption in practice. It requires experts to reach a consensus on the causal graph, the structural equations, the distribution of the input exogenous variables, and to test the validity of their model on available data. This is not a realistic scenario, especially when dealing with a high-number of features and possibly complex structural relations. Besides, this is not practical since a causal model must be designed and tested for each possible dataset. A more straightforward approach is to directly infer the causal model from observational data. There exist for instance sound techniques to learn the causal graph, but they suffer from being NP-hard, with an exponential worst-case complexity with respect to the number of nodes \citep{cooper1990computational,chickering2004large,scutari2019learning}. In addition, this is not enough to compute counterfactual quantities, as the structural equations would still be lacking. To obtain these equations, researchers often predefine the functional form of the relations between the variables on the basis of a known graph (be it assumed or inferred) and learn them through regression models \citep{kusner2017counterfactual,russell2017when}, or infer simultaneously the graph and the structural equations. However, this also becomes computationally challenging as the number of features increases. Notably, the literature mostly addresses simple linear models \citep{shimizu2006linear} or very few variables \citep{hoyer2008nonlinear}. Finally, the approximation error implied by the choice of the functional class can lead to unrealistic, out-of-distribution counterfactuals, as exemplified in Figure~\ref{fig:models} above. To our knowledge, the literature on causal counterfactuals has not pointed out this flaw to date. 

A related issue is causal uncertainty. There exist several causal models corresponding to a same data distribution, leading to possibly different counterfactual models \citep[see][Example 4.2]{bongers2021foundations}. It cannot be tested whether the adjusted model is the \say{true} one, making the modeling inherently uncertain. Moreover, for non-deterministic structural counterfactual models, the computation of counterfactual quantities requires to know the law of the exogenous variables, which is not observable. While it is common to assume a prior distribution on $U$, this also adds uncertainty in the causal modeling, hence on the induced counterfactuals.  

Perhaps more surprisingly, counterfactual quantities are sometimes nonexistent in Pearl's causal framework. The causal modeling we introduced is very general: we do not assume the exogenous variables to be mutually independent, and only suppose that the equations are acyclic. Assumption~\ref{Acyclic} is very common for both practical reasons and reasons of interpretability. In general, however, observational data can be generated through an acyclical mechanism. Critically, (solvable) acyclic models do not always admit solutions under do-interventions, implying that $X_{S=s}$ may not be defined. We refer to \citep[Example 2.17]{bongers2021foundations} for an illustration. As a consequence, counterfactual quantities are ill-defined in such settings.

\subsubsection{Applicability of the transport approach}

Regarding transport-based counterfactual reasoning, the main practical limitation is also computational. The domain $\S$ of the intervened variable $S$ must be finite for the counterfactual model to be tractable. Moreover, generating the model needs $\abs{\S} \left( \abs{\S}-1 \right) / 2$ computations of transportation plans, which can become too expensive when $\abs{\S}$ is large. Therefore, this approach is tailored to settings with small $\abs{\S}$, typically fairness problems where $S$ represents gender or race.

Another inconvenience comes from the fact that one must specify a family of couplings to implement a transport-based counterfactual model. There is no quantitative rule for this choice; it is guided by intuition and feasibility reasons, and we explained above why optimal transport was a relevant option. Note that the causal approach has a similar flaw: as previously explained, structural counterfactual models are subjected to misspecification since the underlying causal model itself is uncertain.
The advantage of transport methods compared to causal modeling is that they circumvent possibly wrong assumptions on the data-generative process. In particular, transport plans consistently adjust to the data (thanks to the marginal constraint) regardless of the chosen family of couplings, whereas misspecification of the SCM may lead to out-of-distribution structural counterfactuals as aforementioned.


In the following, we derive theoretical properties of the counterfactual models introduced in this section, grounding the similarity between optimal transport and Pearl's computation of counterfactuals we evidenced in Figure~\ref{fig:models}. Interestingly, this echoes the work of \cite{black2020fliptest}, who also empirically observed that optimal transport maps generated nearly identical counterfactuals to the ones based on causal models. 

\section{Theoretical results}\label{sec:revisit}

Until now, we have recalled the basics of causality and transport in Sections~\ref{sec:causal} and \ref{sec:transport}, and introduced counterfactual models, either causal-based or transport-based, in Section~\ref{sec:models}. In what follows, we demonstrate connections between both approaches. Concretely, we firstly explore in Section~\ref{sec:depth} the relationship between an SCM and the counterfactual model it induces, providing justifications to what we observed in Section~\ref{sec:law}. More precisely, we study the implications of typical causal assumptions onto the generated counterfactuals. Then, on the basis of these assumptions and the mass-transportation formalism proposed in Section~\ref{sec:models}, we demonstrate in Section~\ref{sec:coincide} that optimal transport recovers structural counterfactuals in specific cases.

\subsection{Causal assumptions and their consequences}\label{sec:depth}

We analyze in detail two standard scenarios of the causal counterfactual framework: first, when the counterfactuals are deterministic---then the computation can be written as an explicit push-forward operation; second, when $S$ can be considered exogenous---then the counterfactual distribution is observable. Note that none of Section~\ref{sec:depth} involves any specific knowledge on optimal transport theory, only on causal modeling and (general) mass transportation.

\subsubsection{The deterministic case}\label{sec:deterministic}

We show that when the SCM deterministically implies the counterfactual values of $X$, then the counterfactual coupling is deterministic. Additionally, we provide the expression of the corresponding push-forward operator. To reformulate structural counterfactuals in deterministic transport terms, we first highlight the functional relation between an instance and its intervened counterparts.
\begin{lemma}\label{lm:functional}
If $\M$ satisfies \ref{Acyclic}, then there exists a measurable function $F$ such that $X \stackrel{\P-a.s.}{=} F(S,U_X)$ and $X_{S=s} \stackrel{\P-a.s.}{=} F(s,U_X)$ for every $s \in S$.
\end{lemma}
The proof leverages the acyclicity of the structural equations, which implies that the system of structural equations defining $X$ and $S$ is triangular, enabling to express $X$ solely in terms of $U_X$ and $S$.

Now, let us set for every $s \in \S$ the function $f_s : u \mapsto F(s,u)$ defined $\mathcal{L}(U_X)$-almost everywhere. Using this notation, we can give a simple expression of the possible counterfactual counterparts of any factual instance. In what follows, $\overline{B}$ denotes the closure of any $B \subseteq \R^d$.
\begin{proposition}\label{prop:support} Let $\M$ satisfy \ref{Acyclic}. For any $s,s' \in \S$ and $\mu_s$-almost every $x \in \X_s$, $$\operatorname{supp}\left(\mu_{\langle s'|s \rangle}(\cdot|x)\right) \subseteq \overline{f_{s'} \circ f_s^{-1}\left(\{x\}\right)}.$$
\end{proposition}
As a direct consequence of this proposition, all counterfactual quantities on $X$ with respect to $S$ are uniquely determined when the right term of the inclusion becomes a singleton, therefore when the following assumption holds.
\begin{description}
    \item[Assumption (I)\namedlabel{Invertibility}{\textbf{(I)}}] \textit{The functions $\{f_s\}_{s \in \S}$ are injective.}
\end{description}
While the unique solvability of acyclic models ensures that $(X,S)$ is deterministically determined by $U$, \ref{Invertibility} states that, conversely, $U_X$ is deterministically determined by $\{X=x,S=s\}$. This assumption holds in particular for \emph{additive-noise models}: classical models where the exogenous variables are additive terms of the structural equations, such as in Example~\ref{ex:scm} and Section~\ref{sec:law}.
\begin{example}\label{ex:additive}
An SCM $\M = \langle U, G \rangle$ is an \emph{additive-noise model} if its causal mechanism $G$ has the form
\[
    G(v,u) := \phi(v) + u,
\]
where $\phi : \mathcal{V} \to \mathcal{V}$ is a measurable function. Under \ref{Acyclic}, therefore unique solvability, each endogenous variable $V_k$ is given by
\[
    V_k \aseq \phi_k\left(V_{\text{Endo}(k)}\right) + U_{\text{Exo}(k)},
\]
where $\phi_k : \V_{\operatorname{Endo}(k)} \to \V_k$. Note that the random seed $U$ is fully determined by the value of $V$, meaning that for any $v \in V$ the posterior distribution $\mathcal{L}(U \mid V=v)$ narrows down to a single value. As such, whatever the do-intervention on $V$, the three-step procedure can only generate a deterministic counterfactual quantity.

Note that in our setting, which addresses interventions on a single endogenous variable $S$, satisfying \ref{Invertibility} does not require a fully invertible model between $V = (X,S)$ and $U$ but simply between $X$ and $U_X$ knowing $S=s$. As illustration, consider a partially-additive-noise model (over $X$ only), namely such that $X$ is generated through
\[
    X \aseq \varphi\left(S,X\right) + U_X,
\]
where $\varphi : \S \times \X \to \X$ is a deterministic measurable function; the equation on $S$ does not matter. Assumption~\ref{Acyclic} entails through unique solvability that $X \aseq (I - \varphi(S,\cdot))^{-1}(U_X)$. After identifying $f_s(u) := (I - \varphi(s,\cdot))^{-1}(u)$, we notice that Assumption~\ref{Invertibility} readily holds such that $f^{-1}_s(x) = x - \varphi(s,x)$.

\end{example}
Remark that Assumption \ref{Invertibility} imposes constraints on the variables and their laws to enable a deterministic correspondence between $X$ and $U_X$. In particular, the two random vectors must live in spaces with same cardinal, preventing for instance a continuous $U_X$ with a discrete $X$. Note also that even though it is restrictive, the mainstream literature on causality frequently assumes full invertibility. In particular, most of the causal-discovery frameworks which aim at inferring the structural equations from observational data require invertible models \citep{zhang2006extensions, hoyer2008nonlinear} or even additive ones \citep{shimizu2006linear}. Analogously, the recent research on causal algorithmic recourse generally addresses invertible models in both theory and practice \citep{karimi2021algorithmic,dominguez2022adversarial,von2022fairness}. In Section~\ref{sec:coincide}, we will use the invertibility assumption as an ideal setting to derive theoretical guarantees.

Let us finally turn to the structural counterfactual models. Assumption \ref{Invertibility} implies that all the couplings between the factual and counterfactual distributions are deterministic, as written in the next proposition.


\begin{proposition}\label{prop:oto}

Let $\M$ satisfy \ref{Acyclic}, suppose that \ref{Invertibility} hold, and for any $s,s' \in \S$ set  the mapping $T^*_{\mathsmaller{ \langle s'|s \rangle}} := f_{s'} \circ f_s^{-1} \restr{\X_s}$ defined $\mu_s$-almost everywhere, where {$f_s^{-1} \restr{\X_s}$ denotes the restriction of $f_s^{-1}$ to $\X_s$.} The following properties hold:
\begin{enumerate}
    \item $\mu_{\langle s'|s \rangle}(\cdot|x) = \delta_{T^*_{\mathsmaller{ \langle s'|s \rangle}}(x)}$ for $\mu_s$-almost every $x \in \X_s$;
    \item $\mu_{\langle s'|s \rangle} = {T^*_{\mathsmaller{ \langle s'|s \rangle}}}_\sharp \mu_s$;
    \item $\pi^*_{\langle s'|s \rangle} = (I \times T^*_{\mathsmaller{ \langle s'|s \rangle}})_\sharp \mu_s$.
\end{enumerate}
We say that $T^*_{\mathsmaller{ \langle s'|s \rangle}}$ is a \emph{structural counterfactual operator}, and identify $\T^* := \{T^*_{\langle s'|s \rangle}\}_{s,s'\in \S}$ to the deterministic structural counterfactual model $\Pi^* = \{(I \times T^*_{\langle s'|s \rangle})_\sharp \mu_s\}_{s,s'\in \S}$.
\end{proposition}

Similarly to the structural counterfactual couplings, the operators in $\T^*$ describe the effect of causal interventions on factual distributions. We highlight that they are well-defined without any knowledge on $\mathcal{L}(U)$, meaning that the exogenous variables are not necessary to compute counterfactual quantities under \ref{Invertibility}.

Lastly, remark that we framed \ref{Invertibility} so that it implies that all the counterfactuals instances for \emph{any} changes on $S$ are deterministic, leading to a fully-deterministic counterfactual model.\footnote{In logic terms, this means that the model verifies the conditional excluded middle \citep{stalnaker:1980}.} However, according to Proposition~\ref{prop:support}, it suffices that one $f_s$ be injective for some $s \in \S$ to render all the counterfactual couplings $\{\pi^*_{\langle s'|s \rangle}\}_{s' \in \S}$ deterministic. Therefore, when \ref{Invertibility} does not hold, the structural counterfactual model possibly contains both random and deterministic couplings.

\subsubsection{The exogenous case}\label{sec:exogenous}

We now discuss the counterfactual implications of the position of $S$ in the causal graph. More specifically, we focus on the case where $S$ can be considered as a root node. We will see that this entails that the structural counterfactual model is a transport-based counterfactual model.

Let $\independent$ denote the independence between random variables. The variable $S$ is said to be \emph{exogenous relative to} $X$ \citep{galles1998axiomatic} if the following holds:

\begin{description}
    \item[Assumption (RE)\namedlabel{Exogeneity}{\textbf{(RE)}}] \textit{$U_S \independent U_X$ and $X_{\text{Endo}(S)} = \emptyset$.}
\end{description}

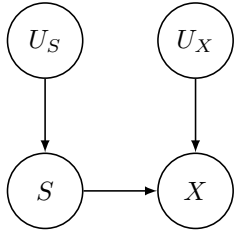
\begin{figure}
    \centering
    \begin{tikzpicture}[-latex ,auto ,node distance =2 cm ,on grid ,
    semithick ,
    state/.style ={circle ,top color =white,
    draw, minimum width =1 cm}]
    \node[state] (S) {$S$};
    \node[state] (X)[right= of S] {$X$};
    \node[state] (Ux)[above=of X]{$U_X$};
    \node[state] (Us)[above=of S]{$U_S$};
    \path (S) edge (X);
    \path (Ux) edge (X);
    \path (Us) edge (S);
    \end{tikzpicture}
    \caption{DAG of a structural causal model satisfying \ref{Exogeneity}. The nodes $U_X$ and $X$ possibly represent several variables.}
    \label{fig:causalmodel}
\end{figure}

The first item, $U_S \independent U_X$, ensures that there is no hidden confounder between $X$ and $S$. The literature on causal modeling generally supposes a stronger condition known as \emph{causal sufficiency}, which states that \emph{all} the $(U_j)_{j \in \J}$ are mutually independent \citep{shimizu2006linear,karimi2021algorithmic,bongers2021foundations,dominguez2022adversarial}. The second item, $X_{\text{Endo}(S)} = \emptyset$, means that $S$ is \emph{ancestrally closed}: no variable in $X$ is a direct cause (or parent) of $S$ (see Figure~\ref{fig:causalmodel}). This holds typically in fairness problems, such as in Section~\ref{sec:law}, where the variable $S$ to alter generally encodes someone's gender, race or age, which do not have any observable causes. As pointed out by \cite{fawkes2021selection}, ancestral closure is a common hypothesis in causal-fairness research, and even a requirement for many frameworks \citep{kusner2017counterfactual,russell2017when,nabi2018fair,chiappa2019path,kilbertus2020sensitivity,plecko2020fair}.

Interestingly, relative exogeneity has critical implications on the generated counterfactuals. Assumption~\ref{Exogeneity} readily entails that $S \independent U_X$. Then, it is easy to see that at the distributional level, intervening on $S$ amounts to conditioning $X$ by a value of $S$.

\begin{proposition}\label{prop:conditioning}

Let $\M$ satisfy \ref{Acyclic}. If \ref{Exogeneity} holds, then for every $s,s' \in \S$ we have $\mu_{S=s'} = \mu_{s'} = \mu_{\langle s'|s \rangle}$.
\end{proposition}

Recall that the structural counterfactual coupling $\pi^*_{ \langle s'|s \rangle}$ represents an intervention transforming an observable distribution $\mu_s$ into an \emph{a priori} non-observable counterfactual distribution $\mu_{\langle s'|s \rangle}$. According to Proposition \ref{prop:conditioning}, \ref{Exogeneity} renders the causal model otiose for the purpose of generating the counterfactual distribution, as the latter coincides with the observable factual distribution $\mu_{s'}$. This is notably what occurred in the example from Section~\ref{sec:law}. However, we underline that the coupling is \emph{still required} to determine how each instance is matched at the individual level. As such, the causal model still carries major information on the induced counterfactual quantities.

Besides, as remarked by \cite{plecko2020fair} and \cite{fawkes2021selection}, a practical consequence of \ref{Exogeneity} is that it enables to link observational and causal notions of fairness. In Section~\ref{sec:fairness}, we will prove a similar result through the prism of counterfactual models. The demonstration relies on the proposition below, which ensures that structural counterfactual models are transport-based counterfactual models when $S$ is relatively exogenous to $X$. 
\begin{proposition}\label{prop:cff}
Let $\M$ satisfy \ref{Acyclic}. If \ref{Exogeneity} holds, then for any $s,s' \in \S$,
\begin{itemize}
    \item[(i)] $\pi^*_{\mathsmaller{ \langle s'|s \rangle}} \in \Pi(\mu_s,\mu_{s'})$;
    \item[(ii)] $t_\sharp \pi^*_{\mathsmaller{ \langle s'|s \rangle}} = \pi^*_{\mathsmaller{ \langle s|s' \rangle}}$.
\end{itemize}
Suppose additionally that \ref{Invertibility} holds. Then, for any $s,s',s'' \in \S$,
\begin{itemize}
    \item[(iii)] ${T^*_{\mathsmaller{ \langle s'|s \rangle}}}_\sharp \mu_s = \mu_{s'}$;
    \item[(iv)] The operator $T^*_{\mathsmaller{ \langle s'|s \rangle}}$ is invertible $\mu_s$-almost everywhere, such that $\mu_{s'}$-almost everywhere ${T^*}^{-1}_{\mathsmaller{ \langle s'|s \rangle}} = T^*_{\mathsmaller{ \langle s|s' \rangle}}$.
\end{itemize}
\end{proposition}
Notably, this means that in classical fairness settings transport-based models can be seen as approximations, relaxations of structural models. Another meaningful consequence of Proposition~\ref{prop:cff} is that items $(ii)$, $(iv)$ and $(v)$ may be false when \ref{Exogeneity} does not hold. Said differently, in general contexts, there is no reciprocity between a factual instance and its structural counterfactual counterparts.

\subsubsection{The example of linear additive SCMs}

We illustrate how our notation and assumptions apply to the case of \emph{linear additive} structural models, which account for many state-of-the-art models \citep{bentler1980linear,shimizu2006linear,hyttinen2012learning,rothenhausler2021anchor}.

\begin{example}\label{ex:linear} Under \ref{Exogeneity} and \ref{Acyclic}, a \emph{linear additive} SCM is characterized by the structural equations
\begin{align*}
    X &= MX + wS + b + U_X,\\
    S &= U_S,
\end{align*}
where $w,b \in \R^d$ and $M \in \R^{d \times d}$ are deterministic parameters such that $I-M$ is invertible, and $U_S \independent U_X$. Solving the equations we get $X = (I-M)^{-1}(wS+b+U_X) =: F(S,U_X)$. Besides, note that \ref{Invertibility} holds such that for any $s \in \S$, $f^{-1}_s(x) = (I-M)x-ws-b$. Then, for any $s,s' \in \S$, $T^*_{\mathsmaller{ \langle s'|s \rangle}}(x) = x + (I-M)^{-1}w(s'-s)$. This general expression is consistent with the example from Section~\ref{sec:law}.

\end{example}

Remarkably, in the specific case of linear additive SCMs fitting \ref{Exogeneity}, computing counterfactual quantities amounts to applying translations between factual distributions. Therefore, should an oracle reveal that the SCM belongs to this class without providing the structural equations, it would suffice to compute the mean translation between sampled points from $\mu_s$ and $\mu_{s'}$ to obtain an estimator of the counterfactual operator $T^*_{\mathsmaller{ \langle s'|s \rangle}}$. For more complex SCMs satisfying \ref{Exogeneity}, it is presumably difficult to infer the counterfactual model from data. We address this issue the next section. Specifically, we show that optimal transport for the quadratic cost generates the same counterfactuals as a class of causal models including linear additive models.

\subsection{When optimal transport meets causality}\label{sec:coincide}

We focus on the deterministic transport-based counterfactual model $\mathcal{T} = \{T_{\mathsmaller{\langle s|s' \rangle}}\}_{s,s' \in \S}$ defined by the solutions of Problem~\eqref{eq:square_monge} between all pairs of factual distributions. That is, for every $s,s' \in \S$,
\begin{equation}\label{eq:our_model}
   T_{\mathsmaller{\langle s'|s \rangle}} := \argmin_{T:\ T_\sharp \mu_s = \mu_{s'}} \int_{\X_s} \norm{x-T(x)}^2 \mathrm{d}\mu_s(x).
\end{equation}
As explained before in Section~\ref{sec:models}, this model provides an elegant interpretation to the obtained counterfactual statements, as they are defined by minimizing the squared Euclidean distance between paired instances, and preserve the quantile between marginals when $d=1$. Moreover, as stated in the following theorem, this transport-based counterfactual model recovers structural counterfactuals in specific cases.

\begin{theorem}\label{thm:monotone}

Let $\M$ satisfy \ref{Acyclic}, \ref{Exogeneity} and \ref{Invertibility}. Suppose that the factual distributions $\{\mu_s\}_{s \in \S}$ are absolutely continuous with respect to the Lebesgue measure and have finite second order moments. If for $s,s' \in \S$, the structural counterfactual operator $T^*_{\mathsmaller{ \langle s'|s \rangle}}$ is the gradient of some convex function, then it is the solution to Problem~\eqref{eq:our_model}.
\end{theorem}
The mass-transportation formalism of Pearl's counterfactual reasoning introduced in Section~\ref{sec:structural} and developed in Section~\ref{sec:depth} renders the proof of this theorem straightforward. The non triviality comes precisely from the reformulation of deterministic structural counterfactuals through push-forward operators. We underline that the demonstration does not require any prior knowledge on optimal transport theory except what we summarized in Lemma \ref{lm:connect}. Thus, for the sake of illustration and clarity, we reproduce it directly below.

\begin{proof}
According to \ref{Invertibility} and Proposition \ref{prop:oto}, the SCM defines a structural counterfactual operator $T^*_{\mathsmaller{ \langle s'|s \rangle}}$ between $\mu_s$ and $\mu_{\langle s'|s \rangle}$. Additionally, \ref{Exogeneity} implies through Proposition \ref{prop:conditioning} that $\mu_{\langle s'|s \rangle} = \mu_{s'}$. Therefore, ${T^*_{\mathsmaller{ \langle s'|s \rangle}}}_\sharp \mu_s = \mu_{s'}$. Assume now that $\mu_s$ is absolutely continuous with respect to the Lebesgue measure, and that both $\mu_s$ and $\mu_{s'}$ have finite second order moments. If $T^*_{\mathsmaller{ \langle s'|s \rangle}}$ is the gradient of some convex function, then according to Lemma \ref{lm:connect} it is the solution to Problem~\eqref{eq:square_monge} between $\mu_s$ and $\mu_{s'}$, that the solution to Problem~\eqref{eq:our_model}.
\end{proof}

Understanding the strengths and limitations of Theorem \ref{thm:monotone} requires understanding how rich is the class of SCMs fitting its assumptions. The larger the class, the more likely optimal transport maps for the squared Euclidean cost will provide (nearly) identical counterfactuals to causality. Finding explicit conditions on $f_s$ and $f_{s'}$ so that $f_{s'} \circ f_s^{-1}$ is the gradient of a convex potential requires tedious computations as soon as $d>1$, which renders the identification of the relevant SCMs difficult. Nevertheless, we can find specific sub-classes of causal models fitting Theorem~\ref{thm:monotone}. For instance, as the structural counterfactual operator from Example~\ref{ex:linear} is the gradient of a convex function, we obtain the following corollary.

\begin{corollary}\label{cor:linear}

Consider a linear additive SCM satisfying \ref{Exogeneity} (see Example \ref{ex:linear}). If the factual distributions $\{\mu_s\}_{s \in \S}$ are absolutely continuous with respect to the Lebesgue measure and have finite second order moments, then for any $s,s' \in \S$, the structural counterfactual operator $T^*_{\mathsmaller{ \langle s'|s \rangle}}$ is the solution to \eqref{eq:square_monge} between $\mu_s$ and $\mu_{s'}$.

\end{corollary}
Therefore, up to a linear approximation of the data-generation process, employing optimal transport maps for counterfactual reasoning in fairness contexts recovers causal changes, as in the example from Section~\ref{sec:law}. Besides, the scope of Theorem \ref{thm:monotone} goes beyond linear additive SCMs, as shown in the following non-linear non-additive example.
\begin{example}
Consider the following SCM,
$$
\begin{cases}
X_1 = \alpha(S) U_1 + \beta_1(S),\\
X_2 = - \alpha(S) \ln^2\left(\frac{X_1-\beta_1(S)}{\alpha(S)}\right) U_2 + \beta_2(S),\\
S = U_S,
\end{cases}
$$
where $\alpha, \beta_1,\beta_2$ are $\R$-valued functions such that $\alpha > 0$, $U_1 > 0$, and $U_S \independent (U_1,U_2)$. It satisfies \ref{Acyclic}, \ref{Invertibility} and \ref{Exogeneity}, such that for any $s,s' \in \S$, the associated structural counterfactual operator is given by,
$$
T^*_{\mathsmaller{ \langle s'|s \rangle}}(x) = \frac{\alpha(s')}{\alpha(s)} x + \left[\beta(s')-\beta(s)\right],
$$
where $\beta = (\beta_1,\beta_2)$ is $\R^2$-valued. This is the gradient of the convex function $x \mapsto \frac{\alpha(s')}{2 \alpha(s)} \norm{x}^2 + \left[\beta(s')-\beta(s)\right]^T x$. Then, if the factual distributions are absolutely continuous with respect to the Lebesgue measure and have finite second-order moments, $T^*_{\mathsmaller{ \langle s'|s \rangle}}$ is the solution to \eqref{eq:square_monge} between $\mu_s$ and $\mu_{s'}$.
\end{example}

Note that the converse of the implication in Theorem \ref{thm:monotone} does not hold. This comes from the fact that many functions (even continuous ones) cannot be written as gradients when $d>1$, as illustrated in the following example.

\begin{example}
Consider the following SCM,
$$
\begin{cases}
X_1 = U_1,\\
X_2 = S X^2_1 + U_2,\\
S = U_S,
\end{cases}
$$
where $U_S \independent (U_1,U_2)$. It satisfies \ref{Acyclic}, \ref{Invertibility} and \ref{Exogeneity}, such that for any $s,s' \in \S$, the associated structural counterfactual operator is given by,
$$
T^*_{\mathsmaller{ \langle s'|s \rangle}}(x_1,x_2) = \left(
x_1,\ x_2 + (s'-s) x^2_1\right).
$$
It cannot be written as the gradient of a function. Consequently, it is not a solution to \eqref{eq:square_monge}.
\end{example}

Through Section~\ref{sec:revisit}, we aimed notably at justifying the pertinence of optimal transport in counterfactual frameworks on top of the insights and illustrations given in Section~\ref{sec:models}. To sum-up, the main requisite for transport-based methods, typically optimal transport, to be used as substitutes for causal counterfactual reasoning is Assumption~\ref{Exogeneity}, ensuring that structural counterfactual models are transport-based counterfactual models. As previously explained, this condition is almost systematically verified in fairness problems, making the proposed surrogate approach relevant in various essential tasks. The more specific assumptions from Theorem~\ref{thm:monotone}, which include \ref{Invertibility}, describe an ideal setting meant to derive theoretical guarantees; optimal transport remains an arguably relevant alternative even outside this context. Altogether, Theorem~\ref{thm:monotone} and Corollary~\ref{cor:linear} support the intuition that computing a $\Pi$ from optimal transport provides a suitable approximation of the unknown structural $\Pi^*$. In the sequel, we apply this approach by extending causal counterfactual frameworks for fairness to transport-based models.

\section{Transport-based counterfactual fairness}\label{sec:fairness}

The strength of the unified mass-transportation viewpoint of counterfactual reasoning we proposed in Section~\ref{sec:models} and further studied in Section~\ref{sec:revisit} lies in the fact that all definitions and frameworks implicitly based on a structural counterfactual model have a transport-based analogue, and can therefore be made feasible. In this section, we apply this process to fairness in machine learning.

Suppose that the random variable $S$ encodes a so-called \emph{sensitive} or \emph{protected attribute} (for example race or gender) which divides the population into different classes in a machine-learning prediction task. We denote by $h : \X \times \S \mapsto \R$ an arbitrary predictor defining the random variable of predicted output $\hat{Y} := h(X,S)$. Fairness addresses the question of the dependence of $\hat{Y}$ on the protected attribute $S$. The most classical fairness criterion is the so-called \emph{demographic} or \emph{statistical parity}, which is achieved when
$\hat{Y} \independent S$.

However, this criterion is notoriously limited, as it only gives a notion of \emph{group fairness}, and does not control discrimination at a subgroup or an individual level: a conflict illustrated by \cite{dwork2012fairness}. The counterfactual framework, by capturing the structural or statistical links between the features and the protected attribute, allows for sharper notions of fairness. We first use the mass transportation formalism introduced in Section \ref{sec:models} to reformulate the accepted \textit{counterfactual fairness} condition \citep{kusner2017counterfactual}. On the basis of the reformulation, we then propose new fairness criteria derived from transport-based counterfactual models. 

\subsection{Causal counterfactual fairness from a mass-transportation viewpoint}

Counterfactual fairness is achieved when individuals and their structural counterfactual counterparts are treated equally.

\begin{definition}\label{def:cf} Let $\M$ satisfy \ref{Acyclic}. A predictor $\hat{Y}=h(X,S)$ is \emph{counterfactually fair} if for every $s,s' \in \S$ and $\mu_s$-almost every $x$ in $\X_s$,

$$
   \mathcal{L}\left(\hat{Y}_{S=s} \mid X=x,S=s\right) = \mathcal{L}\left(\hat{Y}_{S=s'} \mid X=x,S=s\right),
$$
where $\hat{Y}_{S=s} := h(X_{S=s},s)$.

\end{definition}

However, the above definition does not clearly emphasize the pairing between factual and counterfactual values. Interestingly, the mass-transportation viewpoint allows for pair-wise characterizations of counterfactual fairness.

\begin{proposition}\label{prop:rcf}

Let $\M$ satisfy \ref{Acyclic}.

\begin{enumerate}
    \item A predictor $h(X,S)$ is counterfactually fair if and only if for every $s,s' \in \S$ and $\pi^*_{ \langle s'|s \rangle}$-almost every $(x,x')$,
    $$
    h(x,s) = h(x',s').
    $$
    \item If \ref{Exogeneity} holds, then a predictor $h(X,S)$ is counterfactually fair if and only if for every $s,s' \in \S$ such that $s < s'$ and $\pi^*_{ \langle s'|s \rangle}$-almost every $(x,x')$,
    $$
    h(x,s) = h(x',s').
    $$
    \item If \ref{Invertibility} holds, then a predictor $h(X,S)$ is counterfactually fair if and only if for every $s,s' \in \S$ and $\mu_s$-almost every $x$,
    $$
    h(x,s) = h(T^*_{\mathsmaller{ \langle s'|s \rangle}}(x),s').
    $$
    \item If \ref{Invertibility} and \ref{Exogeneity} hold, then a predictor $h(X,S)$ is counterfactually fair if and only if for every $s,s' \in \S$ such that $s<s'$ and $\mu_s$-almost every $x$,
    $$
    h(x,s) = h(T^*_{\mathsmaller{ \langle s'|s \rangle}}(x),s').
    $$
\end{enumerate}

\end{proposition}
Items 2 to 4 in Proposition~\ref{prop:rcf} are variations of the first item under the implications of \ref{Exogeneity} and \ref{Invertibility} through respectively Propositions~\ref{prop:cff}~and~\ref{prop:oto}. Note that they have practical interests. Assumption~\ref{Invertibility} highlights the deterministic relationship between factual and counterfactual quantities and makes unnecessary the knowledge of $\mathcal{L}(U)$ to test counterfactual fairness. Assumption~\ref{Exogeneity} entails by symmetry that only half of the couplings are necessary to check the condition. Additionally, if \ref{Exogeneity} holds, then counterfactual fairness is a stronger criterion than the statistical parity across groups, as shown in the following proposition.

\begin{proposition}\label{prop:stronger} Let $\M$ satisfy \ref{Acyclic} and suppose that \ref{Exogeneity} holds. If the predictor $h(X,S)$ satisfies \textit{counterfactual fairness}, then it satisfies \textit{statistical parity}. The converse does not hold in general.
\end{proposition}

\subsection{Extending counterfactual fairness}

One can think of being counterfactually fair as being invariant to
counterfactual operations with respect to the protected attribute. In order to define SCM-free criteria, we generalize this idea to the models introduced in Section \ref{sec:models}.

\begin{definition}\label{def:tcounter}

\begin{enumerate}
    \item Let $\Pi = \{\pi_{\langle s'|s \rangle}\}_{s,s' \in \S}$ be a (random) transport-based counterfactual model. A predictor $h(X,S)$ is \emph{$\Pi$-counterfactually fair} if for every $s,s' \in \S$ and $\pi_{\langle s'|s \rangle}$-almost every $(x,x')$,
    $$
    h(x,s) = h(x',s').
    $$
    \item Let $\T = \{T_{\langle s'|s \rangle}\}_{s,s' \in \S}$ be a deterministic transport-based counterfactual model. A predictor $h(X,S)$ is \emph{$\T$-counterfactually fair} if for every $s,s' \in \S$ and $\mu_s$-almost every $x$,
    $$
    h(x,s) = h(T_{\mathsmaller{ \langle s'|s \rangle}}(x),s').
    $$
\end{enumerate}

\end{definition}
Note that it follows from the symmetry of the transport-based counterfactual models (see items $(ii)$ and $(iii)$ in Definition~\ref{def:TCFM}) that only half of the couplings are truly required in the above conditions. Besides, because the proof of Proposition \ref{prop:stronger} only relies on the assumption that the couplings have factual distributions for marginals, the following proposition automatically holds.

\begin{proposition}\label{prop:Tcf} Let $\Pi$ be a transport-based counterfactual model (deterministic or not). If a predictor $h(X,S)$ satisfies $\Pi$-counterfactual fairness, then it satisfies statistical parity. The converse does not hold in general.
\end{proposition}

This result has interesting consequences. Consider that, for the purpose of computing counterfactual quantities, some practitioners designed a candidate SCM fitting the data and satisfying \ref{Exogeneity}. Even if the SCM is misspecified, it would still characterize a transport-based counterfactual model controlling statistical parity. The fair data-processing transformation proposed by \cite{plecko2020fair} is an illustrative example.

More generally, the conceptual interest of transport-based fairness criteria is the same as the original counterfactual fairness criterion: they offer notions of individual fairness while still controlling for discrimination against protected groups. The added value is their feasibility. In contrast to Definition \ref{def:cf} and Proposition \ref{prop:rcf}, Definition \ref{def:tcounter} relies on computationally feasible counterfactual models that obviate any assumptions about the data-generation process. In addition, as Definition \ref{def:cf} amounts to $\Pi^*$-counterfactual fairness (when \ref{Exogeneity} holds), one can as well think of Definition \ref{def:tcounter} as an approximation of counterfactual fairness.

Crucially, these new criteria can naturally be applied in classical explainability and fairness machine learning frameworks based on counterfactual reasoning. While \cite{black2020fliptest} focused on explaining discriminatory biases in binary decision rules, we address the training of a $\Pi$-counterfactually fair predictor in Section \ref{sec:application}.

\subsection{Ethical risk}

We conclude this section by discussing a potentially negative impact of our work. As aforementioned, the transport-based approach allows for many counterfactual models, but they do not all define legitimate notions of counterparts. Consequently, transport-based notions of counterfactual fairness could be used for unethical fair-washing. The next proposition formalizes this risk.

\begin{proposition}\label{prop:attack}
If $h(X,S)$ is a classifier satisfying statistical parity, then there exists a transport-based counterfactual model $\Pi$ such that $h(X,S)$ satisfies $\Pi$-counterfactual fairness.
\end{proposition}

Practitioners could take advantage of the weak notion of statistical parity to construct counterfactual models such that their trained classifiers are counterfactually fair, while still discriminating at the subgroup or individual level. This is why we argue that practitioners must always be able to justify the counterfactual models when not imposed by legal experts of the prediction task.

\section{Application to counterfactually fair learning}\label{sec:application}

We now address an application of transport-based counterfactual models to fairness. More precisely, we introduce a supervised learning procedure trading-off between $\Pi$-counterfactual fairness and accuracy, and provide statistical guarantees.

\subsection{Learning problem}\label{sec:learning}

In \citep{russell2017when}, the authors considered a learning problem involving a penalization controlling the pair-wise difference in decision between the training inputs and their structural counterfactual counterparts. While they gave empirical evidence of the efficiency of their training method, they had to assume a known causal model and did not provide consistency guarantees on the estimated predictor. In this sub-section, we illustrate that this counterfactual approach can naturally be made both feasible and statistically consistent by replacing the structural counterparts by transport-based counterparts. Note that in contrast to \cite{russell2017when}, we do not optimize over several counterfactual models.

Let $Y : \Omega \to \Y \subseteq \R$ denote the so-called \emph{ground-truth} variable to predict, and denote by $\D$ the law of the data $(X,S,Y)$. We consider a parametric class of predictors $\{h_\theta\}_{\theta \in \Theta}$ from $\X \times \S$ to $\Y$, indexed by $\Theta \subseteq \R^p$ where $p > 1$. For a given counterfactual model $\Pi := \left\{ \pi_{\langle s' \mid s \rangle} \right\}_{s,s' \in \S}$ and a given weight $\lambda > 0$, we define the following \emph{expected} risk on the predictors,
\begin{equation}\label{expected}
    \mathcal{R}_{\D,\Pi,\lambda}(\theta) :=  \E[\ell(h_\theta(X,S),Y)] + \lambda \sum_{s \in \S} \P(S=s) \sum_{s' \neq s} \E_{}[r_\theta(X_s,s,X_{s'},s')],
\end{equation}
where $\mathcal{L}((X_s,X_{s'})) = \pi_{\langle s' \mid s \rangle}$ for every $s,s' \in \S$. The application $\ell$ denotes a data-loss function, continuous with respect to each of its input variables, while $r_\theta(x,s,x',s')$ is a penalty promoting counterfactual fairness by enforcing the difference between the outputs of the algorithm for an individual and its counterfactual, namely $\abs{h_\theta(x,s)-h_\theta(x',s')}$, to be small. For instance, in \citep{russell2017when}, the authors considered the tightest convex relaxation of \emph{$\epsilon$-approximate counterfactual fairness}, that is $r_\theta(x,s,x',s') := \max\big\{0, \abs{h_\theta(x,s)-h_\theta(x',s')}-\epsilon \big\}$ for some $\epsilon > 0$. In this paper, we rather work with the penalty $r_\theta(x,s,x',s') := \abs{h_\theta(x,s)-h_\theta(x',s')}^2$ which is smoother. Through $\lambda$, the risk $\mathcal{R}_{\D,\Pi,\lambda}$ quantifies a trade-off between accuracy and counterfactual fairness. Importantly, when $\Pi = \Pi^*$, it corresponds precisely to the expected risk of the learning problem proposed by \cite{russell2017when} reframed using the mass-transportation viewpoint. In what follows, we will simply write $\mathcal{R}_{\D,\Pi,\lambda}$ as $\mathcal{R}$.

In practice, we learn a predictor by minimizing an empirical version of $\mathcal{R}$. To this end, we need an empirical counterfactual model. Concretely, consider a training set $\{(x_i,s_i,y_i)\}^n_{i=1}$ composed of $n$ i.i.d. observations drawn from $\D$. We divide this collection into $\S$ protected categories by defining for any $s \in \S$ the index $\I^n_s := \{1 \leq i \leq n \mid s_i = s \}$ of length $n_s := \abs{\I^n_s}$. Then, the empirical versions of the factual distributions are for every $s \in \S$, $\mu^n_s := {n^{-1}_s} \sum_{i \in \I^n_{s}}\delta_{x_i}$. In our case, the counterfactual pairs between $\mu^n_s$ and $\mu^n_{s'}$ are estimated \emph{within} the training dataset through an empirical transport plan $\{\pi^n_{\langle s \mid s' \rangle}(i,j)\}_{i \in \I_s, j \in \I_{s'}}$, typically by solving Problem~\eqref{eq:empirical_kanto} as explained in Section~\ref{sec:estimation}. Then, we define the following \emph{empirical} risk, 
\begin{equation}\label{eq:empirical}
 \mathcal{R}_n(\theta) := \frac{1}{n} \sum^n_{i = 1} \ell(h_\theta(x_i,s_i),y_i) + \lambda n_{s_i} \sum_{s' \neq s_i} \sum_{j \in \I^n_{s'}} \pi^n_{\langle s' \mid s_i \rangle}(i,j) r_\theta(x_i,s_i,x_j,s').
\end{equation}

The learning procedure amounts to carrying out a gradient-descent-based routine to minimize $\mathcal{R}_n$. We underline that this procedure, as the original one from \citep{russell2017when}, is tailored to both regression and multi-class classification. It also works for more than two protected groups, but requires the domain of the sensitive variable to be finite.

\subsection{Consistency}

In this part, we focus on the counterfactual model constructed with quadratic optimal transport, and prove the statistical consistency of the learning procedure. Set a sequence $\left\{ \theta_n \right\}_{n \in \N^*}$ defined by $\theta_n \in \argmin_{\theta \in \Theta} \mathcal{R}_n(\theta)$. The next theorem ensures the convergence to zero of the excess risk $\mathcal{R}(\theta_n) - \min_{\theta \in \Theta} \mathcal{R}(\theta)$ for ball-constrained linear predictions.

\begin{theorem}\label{thm:consistency}


Suppose that for every pair of factual distributions, the Kantorovich problem \eqref{eq:kanto} with cost $c(x,x') := \norm{x-x'}^2$ admits a unique solution. Thus, we can define the counterfactual model $\Pi = \{ \pi_{\langle s' \mid s \rangle} \}_{s,s' \in \S}$ and its empirical counterpart $\Pi^n = \{ \pi^n_{\langle s' \mid s \rangle} \}_{s,s' \in \S}$ as, for every $s,s' \in \S$,

\begin{align}
    \pi_{\langle s' \mid s \rangle} &:= \argmin_{\pi \in \Pi(\mu_s,\mu_{s'})} \int_{\X_s \times \X_{s'}} \norm{x-x'}^2 \mathrm{d} \pi, \label{sol}\\
    \pi^n_{\langle s' \mid s \rangle} &\in \argmin_{\pi \in \Sigma(n_s,n_{s'})} \sum_{i \in \I_s}\sum_{j \in \I_{s'}} \norm{x_i-x_j}^2 \pi(i,j). \label{emp_sol}
\end{align} 

Now, let $\Phi : \X \times \S \to \R^{p}$ be a feature map such that for every $s \in S$ and $x_1,x_2 \in \X$, $\norm{\Phi(x_1,s) - \Phi(x_2,s)} \leq L_s \norm{x_1-x_2}$ where $L_s > 0$. Consider for $\Theta \subseteq \R^p$ the class of linear predictors $\left\{h_\theta\right\}_{\theta \in \Theta}$ defined as $h_\theta(x,s) := \theta^T \Phi(x,s)$. If the following assumptions hold,

\begin{itemize}
    \item[(i)] there exists $D>0$ such that $\Theta = \left\{ \theta \in \R^{p} \mid \norm{\theta} \leq D\right\}$,
    \item[(ii)] there exists $R > 0$ such that $\X \subseteq \left\{ x \in \R^d \mid \norm{x} \leq R\right\}$,
    \item[(iii)] there exists $b > 0$ such that $\Y \subseteq \left\{ y \in \R \mid \abs{y} \leq b\right\}$,
    \item[(iv)] there exists $L>0$ such that for any $(x,s,y) \in \X \times \S \times \Y$, the function $\theta \in \Theta \mapsto \ell(\theta^T \Phi(x,s),y)$ is $L$-Lipschitz,
\end{itemize}
then,

$$
    \mathcal{R}(\theta_n) - \min_{\theta \in \Theta} \mathcal{R}(\theta) \xrightarrow[n \to +\infty]{a.s.} 0.
$$

\end{theorem}
The proof analyzes separately the accuracy term from the regularization term. The demonstration for the former follows classical results from the statistical-learning literature; the demonstration for the latter is original: we firstly show that each penalty contribution can be bounded by a distance between the empirical and the true coupling, and then invoke the convergence in law. We gather some additional remarks below.
\begin{remark}

\begin{enumerate}
    \item Uniqueness of the solution \eqref{sol} holds when the factual distributions are Lebesgue absolutely-continuous, or uniform over a same number of atoms. 
    \item Typically, $\Phi$ is defined as $(x,s) \mapsto (x,s,1)$ in order to add an intercept, or corresponds to the feature map of a kernel when aiming for non-linear decision boundaries.
    \item Assumptions $(i)$ to $(iv)$ are common for supervised learning problems. The sets $\mathcal{X}$ and $\mathcal{Y}$ are usually bounded spaces, as well as $\Theta$ the set of parameters defining the algorithm. The Lipschitz conditions for the loss functions and the feature map can be directly assumed or are direct consequences of smoothness properties and compactness assumptions of the spaces on which they are defined.
    \item The assumption on the second-order moments of the factual distributions is automatically satisfied under $(ii)$. 
    \item If the risks $\mathcal{R}_n$ and $\mathcal{R}$ are strictly convex, then $\theta_n = \argmin_{\theta \in \Theta} \mathcal{R}_n(\theta)$ and $\theta^* = \argmin_{\theta \in \Theta} \mathcal{R}(\theta)$ are well-defined, and it follows that $ {\theta_n} \xrightarrow[n \to +\infty]{a.s.} {\theta^*}$ (this additional step is detailed in the proof of Theorem~\ref{thm:consistency}).
\end{enumerate}

\end{remark}

\section{Numerical experiments} \label{sec:numerical}

In this section, we present the implementation of our counterfactually fair learning procedure on real data, and show that it has the expected behaviour. The code is available at \url{https://github.com/lucasdelara/PI-Fair}.

\subsection{Procedure}

Whatever the dataset, the general procedure is the following: after dividing the studied dataset into a training set and a testing set, we learn one empirical counterfactual models for each set. The first one implements the penalty of the training loss function; the second enables to evaluate the counterfactual fairness of the trained predictors. We compute the corresponding optimal transport plans using the default (non-regularized) POT solver. Then, we train several predictors for various values of the weight $\lambda$ to study the model's ability to trade-off between accuracy and fairness. Finally, we assess the performances of the learnt algorithms according to three criteria: accuracy, group fairness and counterfactual fairness, and we benchmark them against baselines.

\subsubsection{Evaluation metrics}

In what follows, $h : \X \times \S \to \Y$ denotes either a binary classifier ($\Y = \{0,1\}$) or a regression function ($\Y = \R$), and $\D$ denotes a dataset $(X,S,Y)$. Let us properly define the different metrics we employ:

\begin{itemize}
    \item To evaluate the data fidelity of a classifier, we compute the \emph{accuracy} (Acc), defined as
\begin{equation*}
    \operatorname{Acc}(h,\mathcal{D}) := \P(h(X,S)=Y).
\end{equation*}
For a regression function, we compute the \emph{mean square error} (MSE), defined as
\begin{equation*}
    \operatorname{MSE}(h,\mathcal{D}) := \E\left[\norm{h(X,S)-Y}^2\right].
\end{equation*}


    \item To assess the statistical parity of a binary classifier when the protected attribute is binary, we compute the the \emph{parity gap} (PG), defined as
    \begin{equation*}
        \operatorname{PG}(h,\mathcal{D}) := \abs{\P(h(X,S)=1 \mid S=0) - \P(h(X,S)=1 \mid S=1)}.
    \end{equation*}
    It quantifies the violation to group fairness, and equals zero when statistical parity is achieved. For a regression function, we use the \emph{Kolmogorov-Smirnov distance} (KS) between $\mathcal{L}(\hat{Y} \mid S=0)$ and $\mathcal{L}(\hat{Y} \mid S=1)$, defined as
    \begin{equation*}
        \operatorname{KS}(h,\mathcal{D}) := \sup_{y \in \R} \abs{\E[ \mathbf{1}_{\{ h(X,S) > y \}} \mid S=0] - \E[ \mathbf{1}_{\{ h(X,S) > y \}} \mid S=1]}.
    \end{equation*}
    Note that this extends the parity gap to the continuous case. The purpose of these two group-fair indicators is to illustrate Proposition~\ref{prop:Tcf}, stating that counterfactual fairness implies statistical parity.

    \item Finally, we need a metric to evaluate counterfactual fairness. We extend the notion of \emph{$(\epsilon,\delta)$-approximate counterfactual fairness} introduced by \cite{russell2017when} to transport-based counterfactual models. For a counterfactual model $\Pi$ and a tolerance $\epsilon>0$, we define the probability for the disparate treatment by $h$ between $(x,s)$ and its $s'$-counterfactual counterpart to be lower than $\epsilon$ as
    \begin{equation*}
        \operatorname{CFT}_{\epsilon}(h,x,s,s',\Pi) :=\int_{x' \in \X_{s'}} \mathbf{1}_{\left\{\abs{h(x,s)-h(x',s')} \leq \epsilon\right\}} \frac{\mathrm{d}\pi_{\langle s' | s \rangle}}{ \mathrm{d}\mu_s }(x'|x).
    \end{equation*}
    Then, for a probability threshold $0 \leq \delta \leq 1$, we say that a predictor $h$ is \emph{$(\epsilon,\delta)$-approximately counterfactually fair} if for every $s \in \S$, for $\mu_s$-almost every $x \in \X_s$, and for every $s' \neq s$, 
    \begin{equation}\label{eq:CFR}
        \operatorname{CFT}_{\epsilon}(h,x,s,s',\Pi) \geq 1-\delta.
    \end{equation}
    We make two remarks: firstly, if $h$ is a classifier, then the only relevant value for $\epsilon$ is 0; secondly, if the counterfactual model is deterministic, then the only relevant value for $\delta$ is 0. As the empirical counterfactual models we use are non-deterministic---although their continuous counterparts may be deterministic---we set $\delta=0.1$ whatever the prediction task. In practice, we quantify counterfactual fairness through the \emph{$(\epsilon,\delta)$-counterfactual fairness rate} (CFR),
    \begin{equation*}
        \operatorname{CFR}_{\epsilon,\delta}(h,\D,\Pi) := \sum_{s \in \S} \P(S=s) \int_{x \in \X_{s}} \left( \prod_{s' \neq s} \mathbf{1}_{\{\operatorname{CFT}_\epsilon(h,x,s,s',\Pi) \geq 1-\delta\}} \right) \mathrm{d}\mu_s(x).
    \end{equation*}
    This corresponds to the proportion of points satisfying Condition~\ref{eq:CFR}. In the classification setting we set $\epsilon=0$ while in the regression setting we work with $\epsilon = \frac{1}{2} \E\left[\abs{Y-Y'}\right]$ where $Y'$ is an independent copy of $Y$.
\end{itemize}

\subsubsection{Baselines}

We aim at applying our regularized approach for several values of the weight $\lambda$ to study the model's ability to trade-off between accuracy and fairness. For classification tasks, we consider logistic models; for regression tasks, we consider linear regression models. Theses choices will be useful in particular to benchmark our method against the one of \cite{zafar2017fairness}, tailored to such models. For a given $\lambda$, we write \textbf{$\Pi$-Fair($\lambda$)} for the corresponding regularized predictor. We compare the obtained results to three baseline algorithms: the best constant predictor \textbf{Const}, which achieves perfect fairness; the group-fair predictor \textbf{Z} developed by \cite{zafar2017fairness}, which is meant to maximize accuracy under an exact-fairness constraint; the unaltered ($\lambda=0$) predictor \textbf{U}, which is presumably the most accurate but also the most unfair predictor. 

\subsection{Datasets}

We carry out the experiments on four datasets: the first two for classification and the last two for regression. Note that in all the considered settings, the sensitive variable $S$ is binary and relatively exogenous to $X$. Table~\ref{tab:datatset} summarizes information about each dataset after preprocessing.

\begin{table}[t]
    \centering
    \begin{tabular}{|c||c|c|c|c|}
         \hline
         Dataset &  Adult & COMPAS & Law & Crimes \\
         \hline
         \hline
         Task & Classification & Classification & Regression & Regression \\
         \hline
         $S:\ $0/1 & Woman/Man & Black/White & Black/White & Black/Non-black \\
         \hline
         $d$ & 35 & 6 & 2 & 97\\
         \hline
         $n_{train}$ & 32,724 & 4,120 & 13,109 & 1,335\\
         \hline
         $n_{test}$ & 16,118 & 2,030 & 6,458 & 659\\
         \hline
    \end{tabular}
    \caption{Datasets}
    \label{tab:datatset}
\end{table}

\subsubsection{Adult}

The Adult Data Set from the UCI Machine Learning Repository \citep{dua2019} has become a gold reference dataset to evaluate and benchmark fairness frameworks. The \emph{classification} task is to predict whether the income of an individual exceeds 50K USD per year based on census data. Concretely, the dataset contains $n=48,842$ instances with $14$ attributes (numerical and categorical). The ground-truth variable $Y$ equals 1 whenever the incomes exceeds 50K, and 0 otherwise. In this work, we set the sensitive variable $S$ to be the \emph{gender}: $S=0$ stands for \emph{female}, while $S=1$ stands for \emph{male}. The potential sources of algorithmic bias against women have been widely studied by \cite{besse2021survey}. They mainly amount to an under representation of women in the dataset, as well as a high correlation between being a woman and having a lower income. Any standard algorithms, optimizing only for accuracy, are bound to be unfair towards women. Before training any models, we process the data using a one-hot-encoding of the categorical attributes. The processing is the exact same as in \citep{besse2021survey}. This leads to a dataset of dimension $d+1=36$ (without the outcome). We divide it into a training set of size $n_{train} = 32,724$ and a testing set of size $n_{test} = 16,118$.

\subsubsection{COMPAS}

The Correctional Offender Management Profiling for Alternative Sanctions (COMPAS) is an infamous score used by US court officers to assess the risk of criminal recidivism. ProPublica analyzed more than 10,000 of cases from Florida, and concluded that black defendants tended to be predicted riskier than they actually were whereas white defendants were often predicted at lower risk than they were.\footnote{\url{https://www.propublica.org/article/how-we-analyzed-the-compas-recidivism-algorithm}} In this part, we follow \cite{kusner2017counterfactual} and try to predict the risk of recidivism while avoiding discrimination against the race, using the same data. Keeping only black and white defendants, we get $n=6,150$ instances with $d+1 = 7$ attributes such as the number of prior offenses and the type of crime they committed. The ground-truth variable $Y$ equals 1 if the individual recidivated and 0 otherwise. We set the sensitive variable $S$ to be the \emph{race}: $S=0$ stands for \emph{black}, while $S=1$ stands for \emph{white}. Finally, we divide the data into a training set of size $n_{train} = 4,120$ and a testing set of size $n_{test} = 2,030$.

\subsubsection{Law School}

This is the dataset used in Section~\ref{sec:law}, gathering statistics from 163 US law schools and more than 20,000 students. Here again we follow \cite{kusner2017counterfactual}, and try to predict the first-year average grade of individuals $Y$ on the basis of the race (black or white) $S$, the entrance-exam score $X_1$, and the grade-point average before law school $X_2$. All in all, we have $d=2$ features excluding the outcome and the protected attributes, and work with $n_{train} = 13,109$ training entries and $n_{test} = 6,458$ testing entries.

\subsubsection{Communities and crimes}

The Communities and Crimes dataset can also be found in the UCI Machine Learning Repository \citep{dua2019}. It contains socio-economics, law enforcement and crime data from communities across the United States. Similarly to \cite{evgenii2020fair}, we consider the problem of predicting the rate of violent crime per ${10}^5$ of population $Y$ with $S=0$ indicating that at least 50\% of the population is black and $S=1$ otherwise. After processing the 128 numerical and categorical attributes composing the dataset, we obtain $d+1 = 98$ features over $n_{train} = 1,335$ training instances and $n_{testing} = 659$ testing instances.

\subsection{Results}

The regularization weight $\lambda$ takes successively all the values in a grid $\left\{10^{-4},10^{-3.5},\ldots,10^1\right\}$. We repeat the training and evaluation processes of our models together with the baselines across 10 repeats for every datasets. As all learning techniques are deterministic, the randomness of the experiments comes uniquely from the division of the dataset into a training and testing sets. The results are reported in the figures below.

\subsubsection{Trade-off between accuracy and fairness}

\begin{figure}[h]
     \centering
     \begin{subfigure}[b]{0.25\textwidth}
         \centering
         \includegraphics[width=\textwidth]{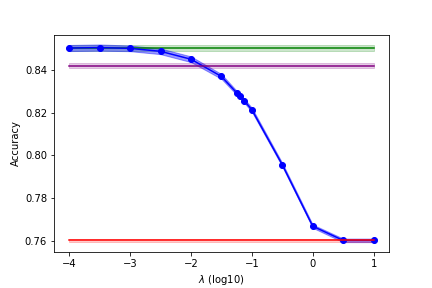}
         \caption{Acc}
         \label{fig:adult_ACC}
     \end{subfigure}
     \begin{subfigure}[b]{0.25\textwidth}
         \centering
         \includegraphics[width=\textwidth]{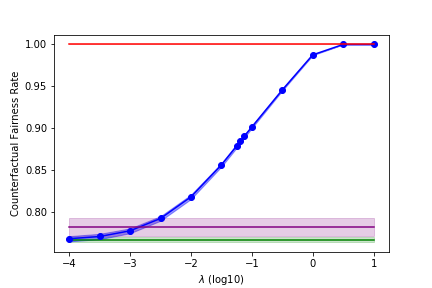}
         \caption{CFR}
         \label{fig:adult_CFR}
     \end{subfigure}
     \begin{subfigure}[b]{0.25\textwidth}
         \centering
         \includegraphics[width=\textwidth]{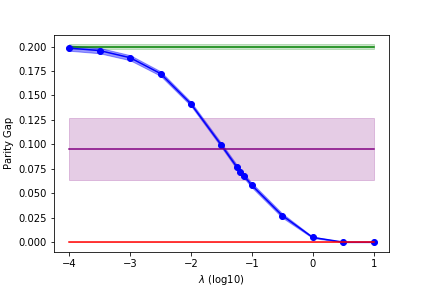}
         \caption{PG}
         \label{fig:adult_DI}
     \end{subfigure}
     \begin{subfigure}[b]{0.20\textwidth}
         \centering
         \includegraphics[width=0.45\textwidth]{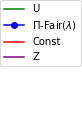}
         \caption{Classifiers}
         \label{fig:adult_legend}
     \end{subfigure}

    \caption{Evaluation metrics on the Adult dataset.}
    \label{fig:adult}
\end{figure}

\begin{figure}[h]
     \centering
     \begin{subfigure}[b]{0.25\textwidth}
         \centering
         \includegraphics[width=\textwidth]{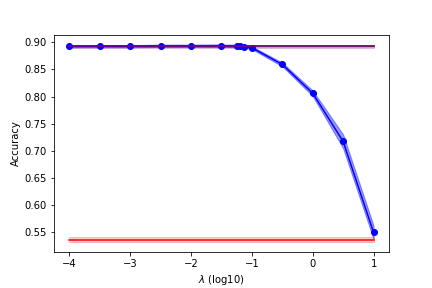}
         \caption{Acc}
         \label{fig:compas_ACC}
     \end{subfigure}
     \hfill
     \begin{subfigure}[b]{0.25\textwidth}
         \centering
         \includegraphics[width=\textwidth]{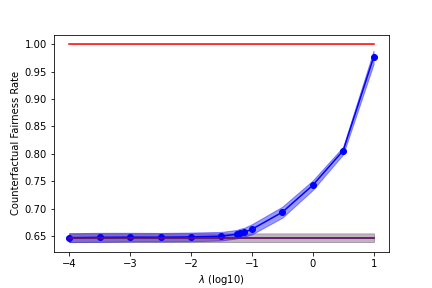}
         \caption{CFR}
         \label{fig:compas_CFR}
     \end{subfigure}
     \begin{subfigure}[b]{0.25\textwidth}
         \centering
         \includegraphics[width=\textwidth]{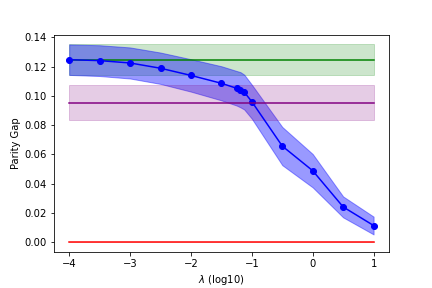}
         \caption{PG}
         
         \label{fig:compas_DI}
     \end{subfigure}
     \begin{subfigure}[b]{0.20\textwidth}
         \centering
         \includegraphics[width=0.45\textwidth]{legend.png}
         \caption{Classifiers}
         \label{fig:compas_legend}
     \end{subfigure}

    \caption{Evaluation metrics on the COMPAS dataset.}
    \label{fig:compas}
\end{figure}

\begin{figure}[h]
     \centering
     \begin{subfigure}[b]{0.25\textwidth}
         \centering
         \includegraphics[width=\textwidth]{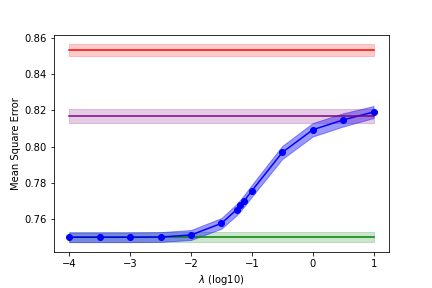}
         \caption{MSE}
         \label{fig:law_MSE}
     \end{subfigure}
     \begin{subfigure}[b]{0.25\textwidth}
         \centering
         \includegraphics[width=\textwidth]{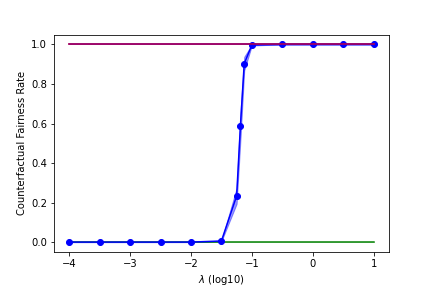}
         \caption{CFR}
         \label{fig:law_CFR}
     \end{subfigure}
     \begin{subfigure}[b]{0.25\textwidth}
         \centering
         \includegraphics[width=\textwidth]{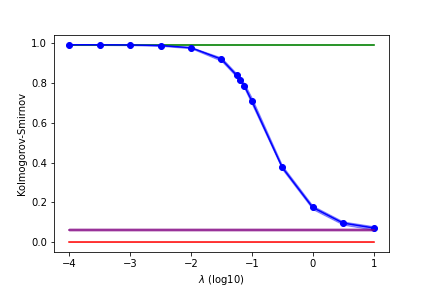}
         \caption{KS}
         \label{fig:law_SP}
     \end{subfigure}
     \begin{subfigure}[b]{0.20\textwidth}
         \centering
         \includegraphics[width=0.45\textwidth]{legend.png}
         \caption{Predictors}
         \label{fig:law_legend}
     \end{subfigure}

    \caption{Evaluation metrics on the Law dataset.}
    \label{fig:law}
\end{figure}

\begin{figure}[h]
     \centering
     \begin{subfigure}[b]{0.25\textwidth}
         \centering
         \includegraphics[width=\textwidth]{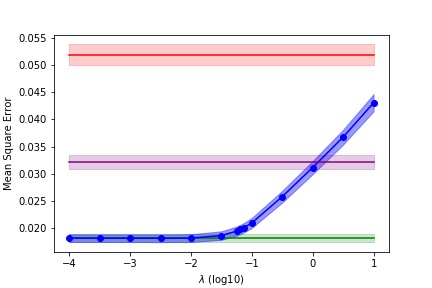}
         \caption{MSE}
         \label{fig:crimes_MSE}
     \end{subfigure}
     \begin{subfigure}[b]{0.25\textwidth}
         \centering
         \includegraphics[width=\textwidth]{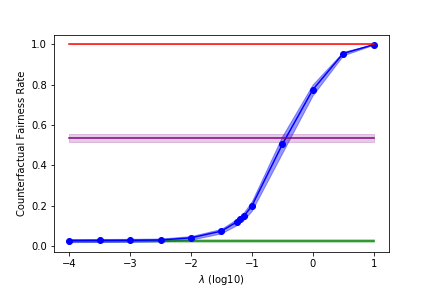}
         \caption{CFR}
         \label{fig:crimes_CFR}
     \end{subfigure}
     \begin{subfigure}[b]{0.25\textwidth}
         \centering
         \includegraphics[width=\textwidth]{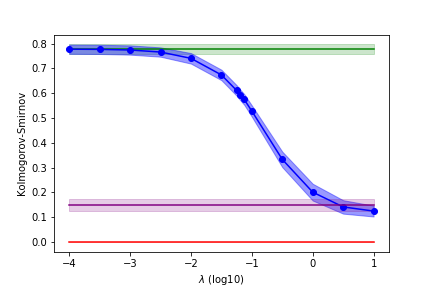}
         \caption{KS}
         \label{fig:crimes_SP}
     \end{subfigure}
     \begin{subfigure}[b]{0.20\textwidth}
         \centering
         \includegraphics[width=0.45\textwidth]{legend.png}
         \caption{Predictors}
         \label{fig:crimes_legend}
     \end{subfigure}

    \caption{Evaluation metrics on the Crimes dataset.}
    \label{fig:crimes}
\end{figure}

Figures~\ref{fig:adult}~to~\ref{fig:crimes} show the evolution with respect to $\lambda$ of the accuracy, the counterfactual fairness rate, and the statistical-parity metric. The solid line represents the mean value of the evaluation metric, while the vertical length of the shaded area corresponds to the standard deviation.

We observe that our learning algorithm is able to reliably trade-off accuracy for counterfactual fairness as $\lambda$ increases, confirming the relevancy of the approach. Additionally, the evaluation metrics remain stable across the different repeats. As anticipated from Proposition~\ref{prop:Tcf}, the regularization also tends to improve group fairness. Overall, the group-fair learning technique of \cite{zafar2017fairness} sacrifices less accuracy than our method to reach the same level of statistical parity, but our method performs better at encouraging counterfactual fairness. We conclude that the prevailing technique depends on the specific type of fairness one wants to achieve. Note that the group-fair predictor \textbf{Z} on the Law dataset (Figure~\ref{fig:law}) behaves similarly to the perfectly counterfactually-fair predictor. This is likely due to the use of simple linear models on such a low-dimensional dataset $(d+1=3)$ limiting the space of feasible algorithms. We leave the in-depth analysis of this phenomenon for further research.



\subsubsection{Recovering causal effects}

\begin{figure}[h]
     \centering
     \begin{subfigure}[b]{0.38\textwidth}
         \centering
         \includegraphics[width=\textwidth]{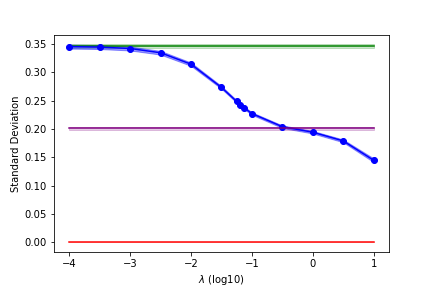}
         \caption{Standard deviation of $\hat{Y}$}
         \label{fig:law_VAR}
     \end{subfigure}
     \begin{subfigure}[b]{0.38\textwidth}
         \centering
         \includegraphics[width=\textwidth]{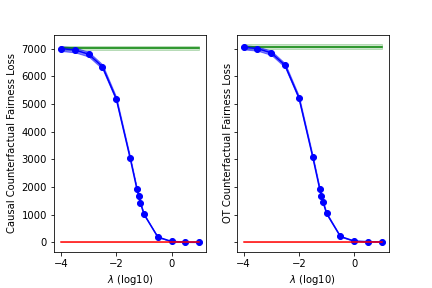}
         \caption{Counterfactual fairness losses}
         \label{fig:law_CCFL}
     \end{subfigure}

    \caption{Promotion of causal counterfactual fairness on the Law dataset.}
    \label{fig:OT_gives_causal}
\end{figure}

To conclude these numerical experiments, let us verify that our optimal-transport counterfactual loss enforces causal counterfactual fairness in the adequate setting. We address the Law dataset for which a plausible causal model is known (see Section~\ref{sec:law}) and satisfies the assumptions of Corollary~\ref{cor:linear}. Figure~\ref{fig:law_CCFL} displays the evolution of the two counterfactual losses, one based on the structural counterfactual model and the other on the optimal-transport counterfactual model, for predictors trained according to the optimal-transport counterfactual model. Figure~\ref{fig:law_VAR} serves as a sanity check: it plots the normalized-variance indicator $\sqrt{\frac{\E\left[\left(\hat{Y}-\E[\hat{Y}]\right)^2\right]}{\E\left[\left(Y-\E[Y]\right)^2\right]}}$ to control how close a predictor is to being constant.

As anticipated by theory, the training process does promote causal counterfactual fairness: the two curves in Figure~\ref{fig:law_CCFL} are almost identical. Crucially, this is not a consequence of the predictors merely becoming constant, since the sequence of predictions in Figure~\ref{fig:law_VAR} have variations that remain significantly higher than the best constant predictor.

\subsection{Discussion}

To sum-up, our learning procedure enables to increase counterfactual fairness while limiting the loss in accuracy, and is both theoretically sound and computationally efficient. This simple approach expands the fair learning arsenal to stronger fairness criteria than group fairness conditions, and so without requiring any additional knowledge on the data-generation process.

Regarding limitations, we note that the current procedure is not tailored to mini-batch learning. Using mini-batches would require to compute a new empirical counterfactual model for each one, which increases the computational complexity, especially since the batch-size should be chosen large enough for the empirical transport plans to make sense. This opens new lines of inquiry for leveraging recent advances on computational optimal transport in order to improve counterfactual learning problems. In particular, we could take advantage of entropic regularization schemes to speed-up the computation of optimal transport plans \citep{cuturi2013sinkhorn,peyre2019computational}. This would make the produced counterfactual model blurry, but still close to the desired solution, allowing a trade-off between precision of the counterfactuals and numerical efficiency. Additionally, we could use the growing literature on plug-in estimations of optimal transport maps \citep{beirlant2020center,hallin2021distribution,manole2021plugin,pooladian2021entropic} to construct empirical counterfactual models not as a matrices, but as a mappings able to generalize to out-of-sample observations, reusable on new datasets and batches. We leave these directions for future work. 

\section{Conclusion}

In this paper, we focused on the challenge of designing sound and feasible counterfactuals. Our work showed that the causal account for counterfactual modeling can be written in a mass-transportation formalism, where implying either deterministic or random counterfactuals has a direct formulation in terms of the deterministic or random nature of couplings between factual and counterfactual instances. This novel perspective enabled us to generalize sharp but unfeasible causal criteria of fairness by actionable transport-based ones. We illustrated that the use optimal transport was a competitive approach to implement these criteria, as it can recover causal changes and can be computed efficiently. In particular, we proposed an new easy-to-implement method to train accurate classifiers with a counterfactual fairness regularization. We provided statistical guarantees, and showed empirically the relevancy of our method. In doing this article, we hope to shed a new light on counterfactual reasoning, and to open lines for strengthening the explainability and fair-learning arsenal in artificial intelligence.


\section*{Acknowledgements}

The authors thank the AI interdisciplinary institute ANITI, grant agreement ANR-19-PI3A-0004 under the French investing for the future PIA3 program.



\newpage

\appendix

\section{Proofs of Section~\ref{sec:revisit}}

\begin{proof}\textbf{of Lemma \ref{lm:functional}}\mbox{ } As a direct consequence of Assumption~\ref{Acyclic}, there exists a topological ordering on the nodes of the graph induced by $\M$. Therefore, starting with the components $X_k$ for which $\operatorname{Endo}(k) = \emptyset$ or $\operatorname{Endo}(k) = \{S\}$, we can recursively replace the terms $X_{\operatorname{Endo}(k)}$ in the formulas $G_k(X_{\operatorname{Endo}(k)},S_{\operatorname{Endo}(k)},U_{\operatorname{Exo}(k)})$ by expressions depending only on $U_X$ and $S$. This yields a measurable function $F$ such that $X \aseq F(S,U_X)$. The same computation but changing $S$ to $s$ for some $s \in \S$ leads to $X_{S=s} \aseq F(s,U_X)$.

\end{proof}

\begin{proof}\textbf{of Proposition \ref{prop:support}}\mbox{ }
Recall that $X \aseq F(S,U_X)$. This implies that, $\P$-almost surely, $(X=x,S=s) \implies U_X \in f_s^{-1}\left(\{x\}\right)$ . Besides, $X_{S=s'} \aseq f_{s'}(U_X)$ according to Lemma~\ref{lm:functional}. Then, let $B \subseteq \X$ be an arbitrary measurable subset and compute:

\begin{align*}
    \P\left(X_{S=s'} \in B \mid X=x,S=s\right)
    &= \P\left(f_{s'}(U_X) \in B \mid X=x,S=s\right)\\
    &= \P\left(f_{s'}(U_X) \in B, U_X \in f_s^{-1}\left(\{x\}\right) \mid X=x,S=s\right)\\
    &= \P\left(f_{s'}(U_X) \in B, f_{s'}(U_X) \in f_{s'} \circ f_s^{-1}(\{x\}) \mid X=x,S=s\right) \\
    &= \P\left(X_{S=s'} \in \left[ B \cap f_{s'} \circ f_s^{-1}\left(\{x\}\right)\right] \mid X=x,S=s\right).
\end{align*}
Therefore, $\mathcal{L}(X_{S=s'} \mid X=x,S=s)$ does not put mass outside $f_{s'} \circ f_s^{-1}(\{x\})$. The definition of the support---the set of points $x \in \R^d$ such that every open neighborhood of $x$ has a positive probability---thus implies that $\operatorname{supp}(\mu_{\langle s'|s \rangle}(\cdot | x)) \subseteq \overline{f_{s'} \circ f_s^{-1}(\{x\})}$.
\end{proof}

\begin{proof}\textbf{of Proposition \ref{prop:oto}}\mbox{ }Set $s,s' \in \S$ and $x \in \X_s$. Note that, according to \ref{Invertibility}, $U_X \aseq f^{-1}_S(X)$. Let us address each item of the proposition separately.

\paragraph{$\bullet$ Item~1.} Proposition~\ref{prop:support} states that $\operatorname{supp}(\mu_{\langle s'|s \rangle}(\cdot | x)) \subseteq \overline{f_{s'} \circ f_s^{-1}(\{x\})}$ for $\mu_s$-almost every $x \in \X_s$. This means according to \ref{Invertibility} that $\operatorname{supp}(\mu_{\langle s'|s \rangle}(\cdot | x)) \subseteq \{f_{s'} \circ f_s^{-1}(x)\}$. Since the support of a probability distribution cannot be empty, we have equality. This proves the first item.

\paragraph{$\bullet$ Item~2.} By definition of the counterfactual distribution, we find that

\begin{align*}
    \mu_{\langle s'|s \rangle} &= \mathcal{L}(X_{S=s'}\mid S=s)\\
    &= \mathcal{L}\left(f_{s'}(U_X)\mid S=s\right)\\
    &= \mathcal{L}\left(f_{s'}\circ f^{-1}_S(X)\mid S=s\right)\\
    &= \mathcal{L}\left(f_{s'}\circ f^{-1}_s(X)\mid S=s\right)\\
    &= \left(f_{s'} \circ f^{-1}_s\right)_\sharp \mu_s.\\
\end{align*}
This proves the second item.

\paragraph{$\bullet$ Item~3.} Similarly, by definition of the structural counterfactual coupling we obtain

\begin{align*}
    \pi_{\langle s'|s \rangle} &= \mathcal{L}\big((X,X_{S=s'})\mid S=s\big)\\
    &= \mathcal{L}\left((X,f_{s'}\left(U_X)\right)\mid S=s\right)\\
    &= \mathcal{L}\left((X,f_{s'}\left(f_s^{-1}\left(X)\right)\right)\mid S=s\right)\\
    &= \mathcal{L}\left((X_s,f_{s'} \circ f_s^{-1}(X_s))\right),
\end{align*}
where $\mathcal{L}(X_s) = \mu_s$. This completes the proof.

\end{proof}

\begin{proof}\textbf{of Proposition \ref{prop:conditioning}}\mbox{ } Set $s \in \S$ and recall that $X \aseq F(S,U_X)$ while $X_{S=s} \aseq F(s,U_X)$. Thanks to Assumption \ref{Exogeneity}, we have that $S \independent U_X$. Therefore,

\begin{align*}
    \mathcal{L}(X\mid S=s) &= \mathcal{L}\left(F(S,U_X)\mid S=s\right),\\   &=\mathcal{L}\left(F(s,U_X)\mid S=s\right),\\
    &=\mathcal{L}\left(F(s,U_X)\right),\\
    &=\mathcal{L}(X_{S=s}).
\end{align*}
This means that $\mu_s = \mu_{S=s}$. Similarly, for $s,s' \in \S$ the counterfactual distribution becomes

\begin{align*}
    \mathcal{L}(X_{S=s'}\mid S=s) &= \mathcal{L}\left(F(s',U_X)\mid S=s\right),\\   &=\mathcal{L}\left(F(s',U_X)\right),\\
    &=\mathcal{L}\left(F(s',U_X)\mid S=s'\right),\\
    &=\mathcal{L}\left(F(S,U_X)\mid S=s'\right),\\
    &=\mathcal{L}(X \mid S=s').
\end{align*}
This means that $\mu_{\mathsmaller{ \langle s'|s \rangle}} = \mu_{s'}$, which completes the proof.

\end{proof}

\begin{proof}\textbf{of Proposition \ref{prop:cff}}\mbox{ } We address each item separately.

\paragraph{$\bullet$ Item~$(i)$.} It is a direct consequence of $\pi^*_{ \langle s'|s \rangle} \in \Pi(\mu_s, \mu_{\langle s'|s \rangle})$ by definition and $\mu_{\langle s'|s \rangle} = \mu_{s'}$ from Proposition~\ref{prop:conditioning}.

\paragraph{$\bullet$ Item~$(ii)$.} Recall that \ref{Exogeneity} implies that $S \independent U_X$. Then, by definition we have
\begin{align*}
    \pi^*_{ \langle s|s' \rangle} &= \mathcal{L}( (X,X_{S=s}) \mid S=s' )\\
    &= \mathcal{L}( (f_{s'}(U_X),f_{s}(U_X)) \mid S=s' )\\
    &= \mathcal{L}( (f_{s'}(U_X),f_{s}(U_X)) \mid S=s )\\
    &= \mathcal{L}( (X_{S=s'},X) \mid S=s )\\
    &= t_\sharp \mathcal{L}( (X,X_{S=s'}) \mid S=s )\\
    &= t_\sharp \pi^*_{ \langle s'|s \rangle}.
\end{align*}

\paragraph{$\bullet$ Item~$(iii)$.} It is a direct consequence of ${T^*_{\mathsmaller{ \langle s'|s \rangle}}}_\sharp \mu_s = \mu_{\langle s'|s \rangle}$ from Proposition~\ref{prop:oto} and $\mu_{\langle s'|s \rangle} = \mu_{s'}$ from Proposition~\ref{prop:conditioning}.


\paragraph{$\bullet$ Item~$(iv)$.} 

We know according to Lemma~\ref{lm:functional} that $X_{S=s} \aseq f_s(U_X)$ and $X_{S=s'} \aseq f_{s'}(U_X)$. Furthermore, it follows from \ref{Exogeneity} and Proposition~\ref{prop:conditioning} that $\mu_{s} = \mathcal{L}(X_{S=s})$ and $\mu_{s'} = \mathcal{L}(X_{S=s'})$. Wrapping this up, there exists a measurable set $\Omega_0 \subseteq \Omega$ with $\P(\Omega_0)=1$ such that for every $\omega \in \Omega_0$, 

\begin{align*}
    X_{S=s}(\omega) &= f_s(U_X(\omega)) \in \X_s,\\
    X_{S=s'}(\omega) &= f_{s'}(U_X(\omega)) \in \X_{s'}.
\end{align*}

In the rest of the proof we implicitly work on an $\omega \in \Omega_0$. Assumption~\ref{Invertibility} ensures that $U_X = f^{-1}_s(X_{S=s})$ so that $X_{S=s'} = \left(f_{s'} \circ f^{-1}_s\right)(X_{S=s})$. Noting that $X_{S=s} \in \X_s$, we
obtain $X_{S=s'} = \left(f_{s'} \circ f^{-1}_s\restr{\X_s}\right)(X_{S=s}) = T^*_{\mathsmaller{ \langle s'|s \rangle}}(X_{S=s})$. Following the same computation after switching $s$ and $s'$, we additionally get that $X_{S=s} = \left(f_{s} \circ f^{-1}_{s'}\restr{\X_{s'}}\right)(X_{S=s'}) = T^*_{\mathsmaller{ \langle s|s' \rangle}}(X_{S=s'})$.

Therefore, ${T^*}_{\mathsmaller{ \langle s'|s \rangle}}$ is invertible on $X_{S=s'}(\Omega_0)$ such that ${T^*}^{-1}_{\mathsmaller{ \langle s'|s \rangle}} = {T^*}_{\mathsmaller{ \langle s|s' \rangle}}$ on $X_{S=s}(\Omega_0)$. Since $\mu_{s}\left(X_{S=s}(\Omega_0)\right)=\P(\Omega_0)=1$ and $\mu_{s'}\left(X_{S=s'}(\Omega_0)\right)=\P(\Omega_0)=1$, this means that ${T^*}_{\mathsmaller{ \langle s'|s \rangle}}$ is invertible $\mu_s$-almost everywhere such that ${T^*}^{-1}_{\mathsmaller{ \langle s'|s \rangle}} = {T^*}_{\mathsmaller{ \langle s|s' \rangle}}$ $\mu_{s'}$-almost everywhere. This completes the proof.

\end{proof}

\begin{proof}\textbf{of Corollary \ref{cor:linear}}\mbox{ } We address the structural equations

\begin{align*}
    X &= MX +wS+ b + U_X,\\
    S &= U_S,
\end{align*}
where $w,b \in \R^d$ and $M \in \R^{d \times d}$ are deterministic parameters. We showed that for any $s,s' \in \S$,
$$
    T^*_{\mathsmaller{ \langle s'|s \rangle}}(x) = x + (I-M)^{-1}w(s'-s).
$$
Notice that $T^*_{\mathsmaller{ \langle s'|s \rangle}}$ is the gradient of the convex function $x \mapsto \frac{1}{2}\norm{x}^2 + \left[(I-M)^{-1}w(s'-s)\right]^T x.$ As \ref{Exogeneity} holds and $\mu_s$ is Lebesgue-absolutely continuous with finite second order moment, it follows from Theorem~\ref{thm:monotone} that $T^*_{\mathsmaller{ \langle s'|s \rangle}}$ is the solution to \eqref{eq:square_monge} between $\mu_s$ and $\mu_{s'}$.

\end{proof}

\section{Proofs of Section \ref{sec:fairness}}

\begin{proof}\textbf{of Proposition \ref{prop:rcf}}\mbox{ } We address each item separately.

\paragraph{$\bullet$ Item~1.} We claim that counterfactual fairness is equivalent to

\begin{description}
    \item[(Goal)\namedlabel{Goal}{\textbf{(Goal)}}] {\it For every $s,s' \in \S$, there exists a measurable set $C = C(s,s') \subseteq \X \times \X$ satisfying $\pi^*_{\langle s'|s \rangle}(C) = 1$ such that for every $(x,x') \in C$
$$
    h(x,s) = h(x',s').
$$}
\end{description}

Note that a direct reformulation of the original counterfactual fairness condition is

\begin{description}
    \item[(CF)\namedlabel{CF}{\textbf{(CF)}}] {\it For every $s,s' \in \S$, there exists a measurable set $A = A(s)$ satisfying $\mu_s(A)=1$, such that for every $x \in A$ and every measurable set $M \subseteq \R$
\begin{equation}\label{cf_eq}
    \P\left(\hat{Y}_{S=s} \in M \mid X=x,S=s\right) =\P\left(\hat{Y}_{S=s'} \in M \mid X=x,S=s\right).
\end{equation}
}
\end{description}

We aim at showing that \ref{CF} is equivalent to \ref{Goal}. To do so, we first prove that one can rewrite \ref{CF} into the following intermediary formulation

\begin{description}
    \item[(IF)\namedlabel{IF}{\textbf{(IF)}}] {\it For every $s,s' \in \S$, there exists a measurable set $A = A(s)$ satisfying $\mu_s(A)=1$, such that for every $x \in A$ and every measurable $M \subseteq \R$ there exists a measurable set $B = B(s,s',x,M)$ satisfying $\mu_{\langle s'|s \rangle}(B|x) = 1$ and such that for every $x' \in B$,
$$
    \mathbf{1}_{\left\{h(x,s)\in M\right\}} = \mathbf{1}_{\left\{h(x',s')\in M\right\}}.
$$}
\end{description}

\paragraph{$\blacktriangleright$ Proof that \ref{CF} $\iff$ \ref{IF}.} Set $s,s',x \in A$ and $M \subseteq \R$. According to the consistency rule, $\mathcal{L}(X \mid S=s) = \mathcal{L}(X_{S=s}\mid S=s)$, we can rewrite the left term of \eqref{cf_eq} as

\begin{align*}
    \P\left(\hat{Y}_{S=s} \in M\mid X=x,S=s\right) &=\P\left(h(X_{S=s},s) \in M\mid X=x,S=s\right)\\
    &= \P(h(X,s),s) \in M\mid X=x,S=s)\\
    &= \P(h(x,s) \in M)\\
    &= \mathbf{1}_{\left\{h(x,s) \in M\right\}}.
\end{align*}

Then, using Definition~\ref{def:3steps}, we reframe the right term of \eqref{cf_eq} as

\begin{align*}
    \P\left(\hat{Y}_{S=s'} \in M\mid X=x,S=s\right) &= \P(h(X_{S=s'},s') \in M\mid X=x,S=s)\\ &= \int \mathbf{1}_{\{h(x',s') \in M\}}\mathrm{d} \mu_{\langle s'\mid s \rangle}(x'|x).
\end{align*}
    
Remark now that because the indicator functions take either the value 0 or 1, the condition $$\mathbf{1}_{\{h(x,s) \in M\}} = \int \mathbf{1}_{\{h(x',s')\in M\}} \mathrm{d} \mu_{\langle s'|s \rangle}(x'|x)$$ is equivalent to $\mathbf{1}_{\{h(x,s)\in M\}} = \mathbf{1}_{\{h(x',s')\in M\}}$ for $\mu_{\langle s'|s \rangle}(\cdot|x)$-almost every $x'$. This means that there exists a measurable set $B = B(s,s',x,M)$ such that $\mu_{\langle s'|s \rangle}(B|x) = 1$ and for every $x' \in B$,
$$
    \mathbf{1}_{\{h(x,s)\in M\}} = \mathbf{1}_{\{h(x',s')\in M\}}.
$$
This proves that \ref{CF} is equivalent to \ref{IF}.

\paragraph{$\blacktriangleright$ Proof that \ref{IF} $\implies$ \ref{Goal}.} As \ref{IF} is true for any arbitrary measurable set $M \subseteq \R$, we can apply this result with $M = \{h(x,s)\}$ to obtain a measurable set $B = B(s,s',x)$ such that $\mu_{\langle s'|s \rangle}(B|x) = 1$ and for every $x' \in B$, $h(x',s') = h(x,s)$. To sum-up, for every $s,s' \in \S$, there exists a measurable set $A = A(s)$ satisfying $\mu_s(A)=1$ such that for every $x \in A$, there exists a measurable set $B = B(s,s',x)$ satisfying $\mu_{\langle s'|s \rangle}(B|x) = 1$, such that for every $x' \in B$, $h(x',s') = h(x,s)$. Now, we must show that the latter equality holds for $\pi^*_{\langle s'|s \rangle}$-almost every $(x,x')$.

To this end, set $C = C(s,s') = \{(x,x') \in \mathcal{X} \times \mathcal{X} | x \in A(s), x' \in B(s,s',x)\}$. Remark that by definition of $A$ and $B$, for every $(x,x') \in C$, $h(x,s) = h(x',s')$. To conclude, let us prove that $\pi^*_{\langle s'|s \rangle}(C) = 1$. 

\begin{align*}
    \pi^*_{\langle s'|s \rangle}(C) &= \int_A \P\left(X_{S=s'} \in B \mid X=x,S=s\right) \mathrm{d} \mu_s(x)\\
    &= \int_A \mu_{\langle s'|s \rangle}(B|x) \mathrm{d} \mu_s(x)\\
    &= \int_A 1 \mathrm{d} \mu_s(x)\\
    &= \mu_s(A)\\
    &= 1.
\end{align*}

This proves that \ref{IF} implies \ref{Goal}.

\paragraph{$\blacktriangleright$ Proof that \ref{Goal} $\implies$ \ref{IF}.} Using \ref{Goal}, consider a measurable set $C = C(s,s')$ satisfying $\pi^*_{\langle s'|s \rangle}(C) = 1$ and such that for every $(x,x') \in C$, $h(x,s) = h(x',s')$. Then, define for any $x \in \mathcal{X}$, the measurable set $B(s,s',x) = \{x' \in \mathcal{X} \mid (x,x') \in C\}$. We use disintegrated formula of $\pi^*_{\langle s'|s \rangle}$ to write

$$
    1 = \int \mu_{\langle s'|s \rangle}(B|x) \mathrm{d} \mu_s(x).
$$

Since $0 \leq \mu_{\langle s'|s \rangle}(B|x) \leq 1$, this implies that for $\mu_s$-almost every $x$, $\mu_{\langle s'|s \rangle}(B|x)=1$. Said differently, there exists a measurable set $A = A(s)$ satisfying $\mu_s(A) = 1$ such that for every $x \in A$, the measurable set $B(s,s',x)$ satisfies $\mu_{\langle s'|s \rangle}(B|x)=1$. By construction of $B$ and by definition of $C$, for every $x \in A$ and every $x' \in B$, $h(x,s) = h(x',s')$. To obtain \ref{IF}, it suffices to take any measurable $M \in \R$ and to note that the latter equality implies that $\mathbf{1}_{\{h(x,s)\in M\}} = \mathbf{1}_{\{h(x',s')\in M\}}$.

\paragraph{$\bullet$ Item~2.} Recall that $\pi^*_{ \langle s|s \rangle} = (I \times I)_\sharp \mu_s$. Therefore, it follows from the previous item that counterfactual fairness can be written as: for every $s,s' \in \S$ such that $s' < s$, and $\pi^*_{ \langle s'|s \rangle}$-almost every $(x,x')$
$$
h(x,s) = h\left(x',s'\right),
$$
and for $\pi^*_{ \langle s|s' \rangle}$-almost every $(x,x')$
$$
h(x,s') = h\left(x',s\right).
$$
Moreover, \ref{Exogeneity} implies through Proposition~\ref{prop:cff} that $\pi^*_{ \langle s|s' \rangle} = t_\sharp \pi^*_{ \langle s'|s \rangle}$. Therefore, the second condition above can be written as: for $\pi^*_{ \langle s'|s \rangle}$-almost every $(x,x')$
$$
h(x',s') = h\left(x,s\right),
$$
which is exactly the first condition. This means that only the first condition is necessary, proving this item.

\paragraph{$\bullet$ Item~3.} Consider \ref{CF}, and recall that for every $s,s' \in S$, $\mu_s$-almost every $x$ and every measurable $M \subseteq \R$ the left term of \eqref{cf_eq} is $\mathbf{1}_{\{h(x,s) \in M\}}$. Let us now reframe the right-term of \eqref{cf_eq}. If \ref{Invertibility} holds, using that $U_X \aseq f_S^{-1}(X)$ we obtain

\begin{align*}
    \P\left(\hat{Y}_{S=s'} \in M \mid X=x,S=s\right)
    &= \P\left(h\left(X_{S=s'},s'\right) \in M \mid X=x,S=s\right)\\
    &= \P\left(h\left(F(s',U_X),s'\right),s') \in M \mid X=x,S=s\right)\\
    &= \P\left(h\left(f_{s'}(f_S^{-1}(X)),s'\right) \in M \mid X=x,S=s\right)\\
    &= \P\left(h\left(f_{s'} \circ f_s^{-1}(x),s'\right) \in M\right)\\
    &= \P\left(h(T^*_{\mathsmaller{ \langle s'|s \rangle}}(x),s') \in M\right)\\
    &= \mathbf{1}_{\left\{h(T^*_{\mathsmaller{ \langle s'|s \rangle}}(x),s') \in M\right\}}.
\end{align*}

Consequently, \ref{CF} holds if and only if, for every measurable $M \in \R$

$$
\mathbf{1}_{\{h(x,s)\in M\}} = \mathbf{1}_{\left\{h(T^*_{\mathsmaller{ \langle s'|s \rangle}}(x),s') \in M\right\}}.
$$

Using the same reasoning as before, we take $M = \{h(x,s)\}$ to prove that this condition is equivalent to $h(x,s) = h(T^*_{\mathsmaller{ \langle s'|s \rangle}}(x),s')$. This concludes the third part of the proof.

\paragraph{$\bullet$ Item~4.} From the previous item and Proposition \ref{prop:conditioning}, it follows that counterfactual fairness can be written as: for every $s,s' \in \S$ such that $s' < s$, for $\mu_s$-almost every $x$

$$
h(x,s) = h\left(T^*_{\mathsmaller{ \langle s'|s \rangle}}(x),s'\right),
$$
and for $\mu_{s'}$-almost every $x$
$$
h(x,s') = h\left({T^*}_{\mathsmaller{ \langle s|s' \rangle}}(x),s'\right).
$$

Set $s,s' \in \S$ such that $s' < s$. To prove the fourth item, we show as for item 2 that the two above
conditions are equivalent. Set $A$ a measurable subset of $\X_s$ such that $\mu_s(A) = 1$, and $h(x,s) = h(T^*_{\mathsmaller{ \langle s'|s \rangle}}(x),s')$ for any $x \in A$. Then, make the change of variable $x' = T^*_{\mathsmaller{ \langle s'|s \rangle}}(x)$ so that $h({T^*}^{-1}_{\mathsmaller{ \langle s'|s \rangle}}(x'),s') = h(x',s')$ for every $x' \in T^*_{\mathsmaller{ \langle s'|s \rangle}}(A)$. By Propositions~\ref{prop:oto} and \ref{prop:conditioning}, ${T^*_{\mathsmaller{ \langle s'|s \rangle}}}_\sharp \mu_s = \mu_{s'}$, which implies that $\mu_{s'}(T^*_{\mathsmaller{ \langle s'|s \rangle}}(A))=1$. Therefore, the equality $h({T^*}^{-1}_{\mathsmaller{ \langle s'|s \rangle}}(x'),s) = h(x',s')$ holds for $\mu_{s'}$-almost every $x'$. Finally, recall that according to Proposition~\ref{prop:cff},  ${T^*}^{-1}_{\mathsmaller{ \langle s'|s \rangle}} = {T^*}_{\mathsmaller{ \langle s|s' \rangle}}$ $\mu_{s'}$-almost everywhere. As the intersection of two sets of probability one is a set of probability one, $h({T^*}_{\mathsmaller{ \langle s|s' \rangle}}(x'),s) = h(x',s')$ holds for $\mu_{s'}$-almost every $x'$. To prove the converse, we can proceed similarly by switching $s$ to $s'$.

\end{proof}

\begin{proof}\textbf{of Proposition \ref{prop:stronger}}\mbox{ } According to Proposition~\ref{prop:rcf}, $h$ is counterfactually fair if and only if for any $s,s' \in \S$ and for $\pi^*_{\langle s'|s \rangle}$-almost every $(x,x')$, $h(x,s) = h(x',s')$ or equivalently $\mathbf{1}_{\{h(x,s)\in M\}} = \mathbf{1}_{\{h(x',s')\in M\}}$ for every measurable $M \in \R$. Set $s,s' \in \S$. Recall that from \ref{Exogeneity}, $\pi^*_{\langle s'|s \rangle}$ admits $\mu_s$ for first marginal and $\mu_{s'}$ for second marginal. Let us integrate this equality with respect to $\pi^*_{\langle s'|s \rangle}$ to obtain, for every measurable $M \subseteq \R$

$$
    \int \mathbf{1}_{\{h(x,s)\in M\}} \mathrm{d} \mu_s(x) = \int \mathbf{1}_{\{h(x',s')\in M\}} \mathrm{d} \mu_{s'}(x).
$$
This can be written as,

$$
    \P(h(X,s) \in M \mid S=s) = \P(h(X,s') \in M \mid S=s'),
$$
which means that
$$
    \mathcal{L}(h(X,S) \mid S=s) = \mathcal{L}(h(X,S) \mid S=s').
$$
As this holds for any $s,s' \in \S$, we have that $h(X,S) \independent S$.

One can easily convince herself that the converse is not true. As a counterexample, consider the following causal model,
$$
    X = S \times U_X + (1-S) \times (1-U_X).
$$
Where $S$ follows an arbitrary law and does not depend on $U_X$. Observe that \ref{Exogeneity} is satisfied so that

\begin{align*}
    \mathcal{L}(X_{S=0}) &= \mathcal{L}(X \mid S=0),\\
    \mathcal{L}(X_{S=1}) &= \mathcal{L}(X \mid S=1),\\
    \mathcal{L}(X \mid S=0) &= \mathcal{L}(X \mid S=1).
\end{align*}

In particular, whatever the chosen predictor, statistical parity will hold since the observational distributions are the same. By definition of the structural counterfactual operator, we have $T^*_{\mathsmaller{ \langle 1|0 \rangle}}(x) = 1-x$.
Now, set the \textit{unaware} predictor (i.e., which does not take the protected attribute as an input), $h(X) := \text{sign}(X-1/2)$. Clearly,

$$
    h(T^*_{\mathsmaller{ \langle 1|0 \rangle}}(x)) = - h(x) \neq h(x).
$$

\end{proof}

\begin{proof}\textbf{of Proposition \ref{prop:attack}}\mbox{ }
Suppose that the classifier $h(X,S)$ takes values in the finite set $\mathcal{Y} \subset \R$, and define for any $s \in \S$ and $y \in \mathcal{Y}$ the sets $\mathcal{H}(s,y) := \{ x \in \R^d \mid h(x,s) = y\}$. Statistical parity can be written as, for any $s \in \S$ and any $y \in \mathcal{Y}$,

$$
\mu_s\left(\mathcal{H}(s,y)\right) = p_y,
$$
where $\{p_y\}_{y \in \mathcal{Y}}$ is a probability on $\mathcal{Y}$ that does not depend on $s$.

Now, set $s,s' \in \S$. We aim at constructing a coupling $\pi_{\langle s' | s \rangle}$ between $\mu_s$ and $\mu_{s'}$ such that,

$$
\pi_{\langle s' | s \rangle}\left(\left\{ (x,x') \in \R^d \times \R^d  \mid  h(x,s)=h(x',s')\right\}\right) = 1.
$$

We define our candidate $\pi_{\langle s' | s \rangle}$ as,

$$
 \mathrm{d} \pi_{\langle s' | s \rangle}(x,x') := \sum_{y \in \mathcal{Y}} \frac{\mathbf{1}_{\left\{x \in \mathcal{H}(s,y)\right\}} \mathbf{1}_{\left\{x' \in \mathcal{H}(s',y)\right\}}}{p_y} \mathrm{d} \mu_s(x) \mathrm{d} \mu_{s'}(x').
$$

First, let's show that it admits respectively $\mu_s$ and $\mu_{s'}$ as first and second marginals. Let $A \subseteq \R^d$ be a measurable set,

\begin{align*}
    \pi_{\langle s' | s \rangle}\left(A \times \R^d\right) &= \sum_{y \in \mathcal{Y}} \int_{\R^d}\int_{A}{\frac{\mathbf{1}_{\{x \in \mathcal{H}(s,y)\}} \mathbf{1}_{\{x' \in \mathcal{H}(s',y)\}}}{p_y} } \mathrm{d} \mu_s(x) \mathrm{d} \mu_{s'}(x')\\
    &= \sum_{y \in \mathcal{Y}} \frac{p_y}{p_y} \int_A  \mathbf{1}_{\left\{x \in \mathcal{H}(s,y)\right\}}  \mathrm{d} \mu_s(x)\\
    &= \sum_{y \in \mathcal{Y}} \mu_s\left(A \cap \mathcal{H}(s,y)\right)\\
    &= \mu_s(A).
\end{align*}
One can follow the same computation for the second marginal. To conclude, compute

\begin{align*}
    &\pi_{\langle s' | s \rangle}\left(\{ (x,x') \in \R^d \times \R^d \mid  h(x,s)=h(x',s')\}\right)\\ &= \pi_{\langle s' | s \rangle}\left(\bigsqcup_{y \in \mathcal{Y}} \mathcal{H}(s,y) \times \mathcal{H}(s',y)\right)\\
    &= \sum_{y \in \mathcal{Y}} \pi_{\langle s' | s \rangle}\left(\mathcal{H}(s,y) \times \mathcal{H}(s',y)\right)\\
    &= \sum_{y \in \mathcal{Y}} \frac{1}{p_y} \int  \mathbf{1}_{\{x \in \mathcal{H}(s,y)\}}  \mathrm{d} \mu_s(x) \int  \mathbf{1}_{\{x \in \mathcal{H}(s',y)\}}  \mathrm{d} \mu_{s'}(x)\\
    &= \sum_{y \in \mathcal{Y}} \frac{1}{p_y} p_y \times p_y\\
    &= 1.
\end{align*}

\end{proof}

\section{Proofs of Section \ref{sec:application}}

\begin{proof}\textbf{of Theorem \ref{thm:consistency}}\mbox{ }
The outline of the proof is typical for such supervised learning problems, though some parts require basic knowledge on optimal transport. It mainly amounts to show the uniform convergence of $\{\mathcal{R}_n\}_{n \in \N^*}$ to $\mathcal{R}$, to then use the following classical deviation inequality,

\begin{equation} \label{deviation}
    \mathcal{R}(\theta_n) - \min_{\theta \in \Theta} \mathcal{R}(\theta) \leq 2 \sup_{\theta \in \Theta} \abs{\mathcal{R}_n(\theta) - \mathcal{R}(\theta)}.
\end{equation}

For any measure $P$ and any measurable function $g$, we will use the notation $P(g) := \int g \mathrm{d} P$ throughout the proof.

\paragraph{$\bullet$ Step 1. Uniform convergence of the risk.} By the triangle inequality,

\begin{align*}
    \sup_{\theta \in \Theta} \abs{\mathcal{R}_n(\theta) - \mathcal{R}(\theta)} \leq & \sup_{\theta \in \Theta} \abs{\frac{1}{n} \sum^n_{i=1} \ell(h_\theta(x_i,s_i),y_i) - \E\left[\ell(h_\theta(X,S),Y)\right]}\\
    & + \lambda \sum_{s \in \S} \sum_{s' \neq s} \sup_{\theta \in \Theta} \abs{\left(\frac{n_s}{n} \pi^n_{\langle s' | s \rangle}- \P(S=s) \pi_{\langle s' | s \rangle}\right)\left(r_\theta(\cdot,s,\cdot,s')\right)}.
\end{align*}

The first term corresponds to the standard uniform risk deviation of supervised learning problems for Lipschitz losses and linear predictions. Under Assumptions $(i)$ to $(iv)$, for $0 < \delta < 1$ it follows from \citep[Theorem 26.5]{shalev2014understanding} that with probability greater than $1 - \delta$,

$$
    \sup_{\theta \in \Theta} \abs{ \frac{1}{n} \sum^n_{i=1} \ell(h_\theta(x_i,s_i),y_i) - \E\left[\ell(h_\theta(X,S),Y)\right]} \leq \frac{\ell_0 + L D}{\sqrt{n}}\left(2 + \sqrt{2 \log \frac{1}{\delta}}\right),
$$
where $\ell_0 = \sup_{\abs{y} \leq b} \abs{\ell(0,y)}$. Then, by taking $\delta_n := \frac{1}{n^2}$, we apply Borel-Cantelli lemma so that for every $\omega \in \Omega$, there exists a threshold $N(\omega)$ such that for any $n \geq N(\omega)$,

$$
    \sup_{\theta \in \Theta} \abs{\frac{1}{n} \sum^n_{i=1} \ell(h_\theta(x_i,s_i),y_i) - \E\left[\ell(h_\theta(X,S),Y)\right]} \leq \frac{\ell_0 + L D}{\sqrt{n}}\left(2 + \sqrt{4 \log n}\right).
$$
The upper bound tends to zero as $n$ tends to infinity, and consequently $$\sup_{\theta \in \Theta} \abs{\frac{1}{n} \sum^n_{i=1} \ell(h_\theta(x_i,s_i),y_i) - \E\left[\ell(h_\theta(X,S),Y)\right]} \xrightarrow[n \to + \infty]{a.s.} 0.$$

The critical part is dealing with the counterfactual penalization. Let $s,s' \in \S$ such that $s' \neq s$. In the following of this step, we aim at showing that,
$$
    \sup_{\theta \in \Theta} \abs{\left(\frac{n_s}{n} \pi^n_{\langle s' | s \rangle}- \P(S=s) \pi_{\langle s' | s \rangle}\right)\left(r_\theta(\cdot,s,\cdot,s')\right)} \xrightarrow[n \to +\infty]{a.s.} 0.
$$
To do so, we use the triangle inequality again, leading to,
\begin{align}
    &\sup_{\theta \in \Theta} \abs{\left(\frac{n_s}{n} \pi^n_{\langle s' | s \rangle}- \P(S=s) \pi_{\langle s' | s \rangle})\big(r_\theta(\cdot,s,\cdot,s')\right) } \nonumber \\
    & \leq \abs{ \frac{n_s}{n} - \P(S=s)} \sup_{\theta \in \Theta} \int r_\theta(x,s,x',s') \mathrm{d} \pi_{\langle s' | s \rangle}(x,x') \label{t1}\\ &+ \P(S=s) \sup_{\theta \in \Theta} \abs{ \int r_\theta(x,s,x',s')  \left(\mathrm{d}\pi^n_{\langle s' | s \rangle}(x,x') -  \mathrm{d} \pi_{\langle s' | s \rangle}(x,x') \right)} \label{t2}.
\end{align}
The terms \eqref{t1} tends to zero almost surely as $n$ increases to infinity. We now turn to the convergence of the term \eqref{t2}.

Firstly, let us show that the functions $\{r_\theta(\cdot,s,\cdot,s')\}_{\theta \in \Theta}$ are uniformly Lipschitz on $\X \times \X$. For any $(x_1,x'_1),(x_2,x'_2) \in \X \times \X$, we have,
\begin{align*}
    \abs{r_{\theta}(x_1,s,x'_1,s') - r_{\theta}(x_2,s,x'_2,s')} &\leq \abs{\theta^T \left(\Phi(x_1,s)-\Phi(x'_1,s') - \Phi(x_2,s) + \Phi(x'_2,s')\right) }^2 \\
    &\leq \abs{\theta^T \left(\Phi(x_1,s)-\Phi(x_2,s)\right)}^2\\ &+ \abs{\theta^T \left(\Phi(x'_1,s')-\Phi(x'_2,s')\right)}^2,\\
    &\leq \norm{\theta}^2 \norm{\Phi(x_1,s)-\Phi(x_2,s)}^2\\ &+ \norm{\theta}^2 \norm{\Phi(x'_1,s')-\Phi(x'_2,s')}^2,\\
    & \leq D^2 \left\{ L^2_s \norm{x_1-x_2}^2 + L^2_{s'} \norm{x'_1 - x'_2}^2 \right\},\\
    & \leq D^2 \max_{s \in \S} L^2_s \norm{ (x_1,x'_1) - (x_2,x'_2)}^2,\\
    & \leq 4 D^2 \max_{s \in \S} L^2_s R^2 \norm{ (x_1,x'_1) - (x_2,x'_2)}.
\end{align*}
Let us set $\Lambda := 4 D^2 (\max_{s \in \S} L_s)^2 R^2$, so that the functions $\{r_\theta(\cdot,s,\cdot,s')\}_{\theta \in \Theta}$ are $\Lambda$-Lipschitz.

Secondly, we know from \citep[Theorem 5.19]{villani2008optimal} that $\pi^n_{\langle s' | s \rangle}$ converges almost-surely weakly to $\pi_{\langle s' | s \rangle}$ as $n_s,n_{s'} \xrightarrow[]{} +\infty$. Moreover, for any $s \in \S$, we have $\frac{n_s}{n} \xrightarrow[n \to +\infty]{} \P(S=s)>0$, hence $n_s \xrightarrow[n \to +\infty]{} +\infty$. As a consequence, $\pi^n_{\langle s' | s \rangle}$ converges almost-surely weakly to $\pi_{\langle s' | s \rangle}$ as $n \xrightarrow[]{} +\infty$. Additionally, since $\X_s \times \X_{s'} \subseteq \X \times \X$, it follows from Assumption $(ii)$ that $\pi_{\langle s' | s \rangle}$ is compactly supported. According to Remark 7.13 in \citep{villani2003topics}, this implies that $W_1(\pi^n_{\langle s' | s \rangle},\pi_{\langle s' | s \rangle}) \xrightarrow[n \to +\infty]{a.s.} 0$, where $W_1$ denotes the Wasserstein-1 distance. Using the dual formulation of the Wasserstein distance, this convergence can be written as,
\begin{equation*}
    W_1(\pi^n_{\langle s' | s \rangle},\pi_{\langle s' | s \rangle}) = \sup_{r \in \operatorname{Lip}_{1}(\X \times \X,\R)} \int r \left(\mathrm{d}\pi^n_{\langle s' | s \rangle} -  \mathrm{d} \pi_{\langle s' | s \rangle} \right) \xrightarrow[n \to +\infty]{a.s.} 0.
\end{equation*}
Noting that for any $\theta \in \Theta$, $r_\theta(\cdot,s,\cdot,s') / \Lambda$ is 1-Lipschitz, we have
\begin{equation*}
    \frac{1}{\Lambda} \sup_{\theta \in \Theta} \abs{ \int r_\theta(x,s,x',s')  \left(\mathrm{d}\pi^n_{\langle s' | s \rangle}(x,x') -  \mathrm{d} \pi_{\langle s' | s \rangle}(x,x') \right)} \leq W_1(\pi^n_{\langle s' | s \rangle},\pi_{\langle s' | s \rangle}) \xrightarrow[n \to +\infty]{a.s.} 0.
\end{equation*}
This entails that the term \eqref{t2} converges almost surely to 0, and completes the proof.

\paragraph{$\bullet$ Step 2. Consistency of the mimimum.} For this additional step, we assume that $\mathcal{R}_n$ and $\mathcal{R}$ have unique minimizers, and we denote $\theta^* := \argmin_{\theta \in \Theta} \mathcal{R}(\theta)$. Note that the sequence $\{\theta_n\}_{n \in \N^*}$ is bounded by $D$, and as such we can extract a sub-sequence $\{\theta_{\sigma(n)}\}_{n \in \N^*}$ converging to some $\theta_\sigma \in \Theta$. Let us prove that $\theta_\sigma = \theta^*$ regardless of the choice of the subsequence $\{\sigma(n)\}_{n \in \N}$. According to the deviation inequality \eqref{deviation} and by continuity of $\mathcal{R}$, we have at the limit,

$$
    \mathcal{R}(\theta_{\sigma}) \leq \mathcal{R}(\theta^*).
$$

This means that $\theta_{\sigma}$ is a minimizer of $\mathcal{R}$. Therefore, by uniqueness, $\theta_\sigma = \theta^*$. This completes the proof, as this implies that,
$$
    \theta_n \xrightarrow[n \to +\infty]{a.s.} \theta^*.
$$

\end{proof}

\bibliographystyle{abbrvnat}
\bibliography{references}

\end{document}